%% file: main.tex
\PassOptionsToPackage{table}{xcolor}
\documentclass[11pt,letterpaper]{applemlr}
\input{preamble}

\begin{document}
\addtocontents{toc}{\protect\setcounter{tocdepth}{-1}}
\maketitle

\input{sections/main}
\label{maintextend}

\clearpage
\bibliography{references}
\bibliographystyle{assets/plainnat}

\clearpage
\begin{center}
{\LARGE\bfseries Supplementary Material\par}
\end{center}
\appendix
\setcounter{table}{0}
\setcounter{figure}{0}
\setcounter{equation}{0}
\renewcommand{\thetable}{S\arabic{table}}
\renewcommand{\thefigure}{S\arabic{figure}}
\renewcommand{\theequation}{S\arabic{equation}}

\input{sections/supplement}

\end{document}

%% file: preamble.tex
\usepackage{amsmath,amssymb,xspace,enumitem,url}
\usepackage{longtable,array,tabularx,alltt,float,needspace}
\usepackage{fix-cm}
\newcolumntype{Y}{>{\raggedright\arraybackslash}X}
\newcolumntype{R}{>{\raggedleft\arraybackslash}X}
\definecolor{kblue}{HTML}{2369A8}
\definecolor{korange}{HTML}{D87525}
\definecolor{kpale}{HTML}{EEF5FA}
\definecolor{titlegray}{HTML}{F2F2F2}
\hypersetup{colorlinks=true,linkcolor=kblue,citecolor=kblue,urlcolor=kblue,
 pdftitle={Robo-Harness K1: Harnessing Robot-Use Agents via Perception Augmentation},
 pdfauthor={Zexi Li, Yehang Zhang, Wenqian Li, Haojian Huang, Chenxu Wang, Chunyang Zhang, Shiyuan Deng, Yangkai Wei, Tianyi Zhang, Ruize Gao, Binghui Xie, Bohan Zhou, Jinghao Li, Kaiwen Zhou, Yifan Chang, James Cheng, Yinchuan Li}}
\newcommand{\method}{{\normalfont\scshape Robo-Harness K1}\xspace}
\newcommand{\tool}[1]{\texttt{#1}}

\titlespacing*{\section}{0pt}{3ex plus .5ex minus .3ex}{1.2ex}
\titlespacing*{\subsection}{0pt}{2.3ex plus .4ex minus .2ex}{.9ex}
\titlespacing*{\subsubsection}{0pt}{1.8ex plus .3ex minus .2ex}{.7ex}
\let\cite\citep

\title{Robo-Harness K1: Harnessing Robot-Use Agents via Perception Augmentation}
\renewcommand\authorformat[2][]{\mbox{\sffamily\bfseries #2\textsuperscript{#1}}}
\author[1,3,*]{Zexi Li}
\author[2,3]{Yehang Zhang}
\author[1,3]{Wenqian Li}
\author[2,3]{Haojian Huang}
\author[3]{Chenxu Wang}
\author[3]{Chunyang Zhang}
\author[3]{Shiyuan Deng}
\author[3]{Yangkai Wei}
\author[3]{Tianyi Zhang}
\author[3]{Ruize Gao}
\author[3]{Binghui Xie}
\author[1,3]{Bohan Zhou}
\author[1,3]{Jinghao Li}
\author[3]{Kaiwen Zhou}
\author[3]{Yifan Chang}
\author[1]{James Cheng}
\author[3,*]{Yinchuan Li}

\newcommand{\projectlinks}{%
  \href{https://robo-harness.github.io/k1/}{\textbf{\raisebox{.30em}{\begin{tikzpicture}[x=1em,y=1em,line width=.65pt,baseline]
    \draw (0,0) circle (.4); \draw (0,0) ellipse (.18 and .4); \draw (-.4,0)--(.4,0);
  \end{tikzpicture}}\; Project Page}}\hspace{1.8em}%
  \href{https://github.com/Robo-Harness/k1}{\textbf{\texttt{</>}\; GitHub}}}

\renewcommand{\mymaketitle}{%
  \begin{tcolorbox}[enhanced,frame hidden,colback=titlegray,arc=12pt,
    left=12pt,right=12pt,top=10pt,bottom=10pt,before skip=0pt,after skip=4pt]
    \setlength{\parindent}{0pt}\setlength{\parskip}{0pt}
    \begin{minipage}[c]{0.80\linewidth}
      \raggedright\sffamily\bfseries\fontsize{20}{24}\selectfont
      Robo-Harness K1: Harnessing Robot-Use Agents via Perception Augmentation
    \end{minipage}\hfill
    \begin{minipage}[c]{0.18\linewidth}
      \includegraphics[width=\linewidth]{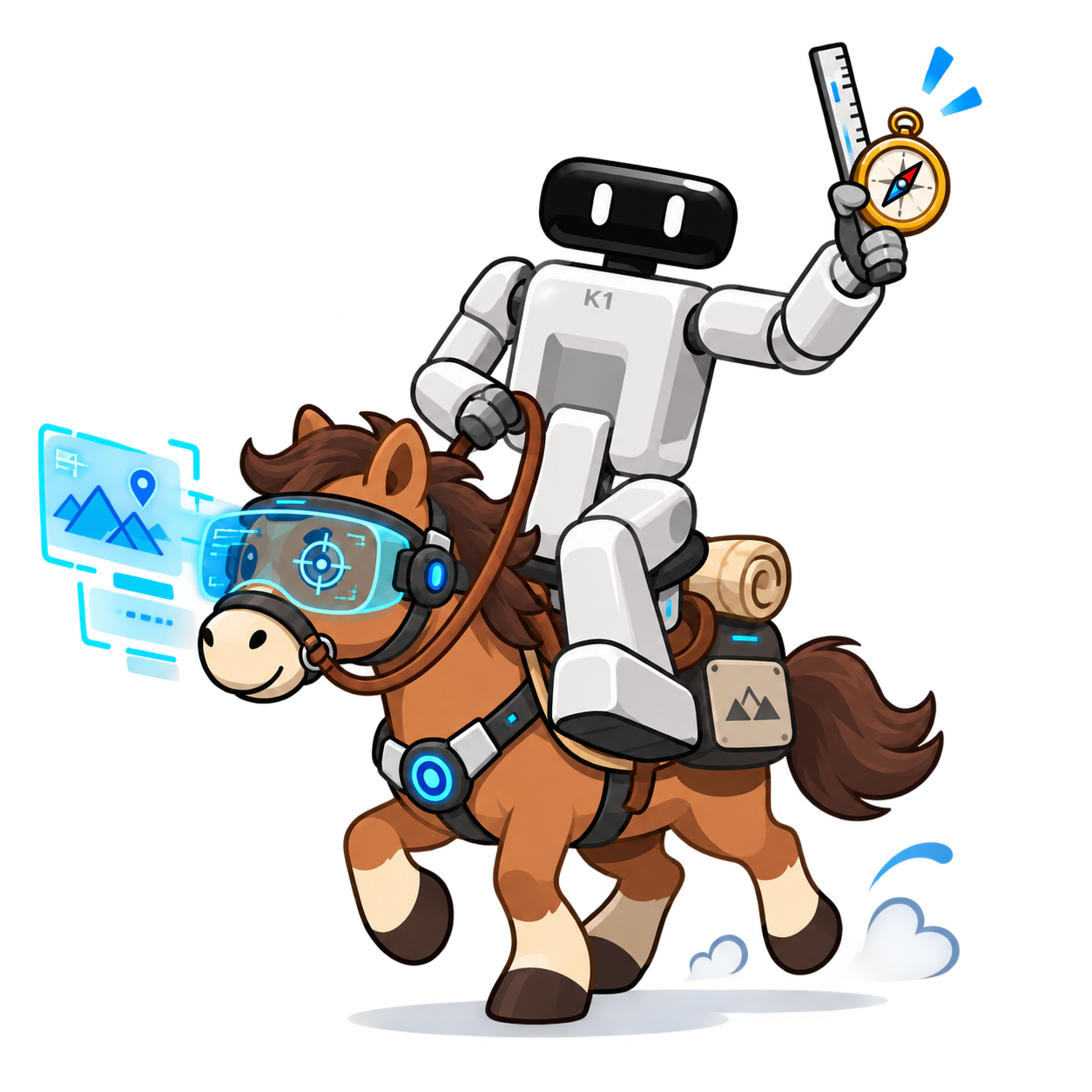}
    \end{minipage}\par\vspace{9pt}
    {\raggedright\fontsize{10.5}{13.2}\selectfont\authorlist\par}
    \vspace{7pt}
    {\raggedright\sffamily\color{fgcolor}\fontsize{9.5}{12}\selectfont
      $^{1}$The Chinese University of Hong Kong\par
      $^{2}$The Hong Kong University of Science and Technology (Guangzhou)\par
      $^{3}$Knowin AI\par
      $^{*}$Corresponding authors:\quad
      \href{mailto:zexili@cuhk.edu.hk}{\texttt{zexili@cuhk.edu.hk}},\quad
      \href{mailto:lizexi@knowin.ai}{\texttt{lizexi@knowin.ai}}\par}
    \vspace{6pt}
    {\sffamily\fontsize{10}{12}\selectfont\projectlinks\par}
    \vspace{6pt}
    {\fontsize{9.5}{11.2}\selectfont\abstractlist\par}
  \end{tcolorbox}
  \FloatBarrier
}

\abstract{Vision-language models (VLMs) understand manipulation scenes and instructions, yet as robot-use agents they struggle to turn 2D observations into the 3D relations that motion requires. We introduce \method, a robot-use agent harness built on \emph{perception as tools}. K1 offers region grounding, calibrated depth and geometry, persistent visual anchors, and projected grasp candidates as callable tools. The VLM calls them to measure the scene, chooses its next motion from the results, and measures again after it acts. On LIBERO-PRO, K1 raises Gemini 3.7 Flash from 16.7\% to 71.7\% and GPT-6 Astra from 63.3\% to 93.3\%, with the largest gains on spatial tasks. Without fine-tuning, the Gemini agent transfers to three RoboSuite arms, dual-arm RoboTwin tasks, and mobile manipulation in RoboCasa. In real-world experiments on a Piper arm, K1 completes eight of nine concluded trials across three tasks. Agentic post-training further distills a frontier model's traces into a compact agent: trained on 107 successful episodes, Qwen3.5-9B reaches 44.2\% on new initial states and 13.9\% on held-out task conditions, versus 30.2\% and 0.0\% for OpenVLA trained on the same episodes. Mixed training on traces from four environments raises held-out accuracy from 11.1\% to 27.8\% and RoboSuite success from 28.3\% to 50.0\%. These results support perceptual tool use for deploying and training robot-use agents.}

\newcommand{\konealiaslabel}[2]{%
  \ifcsdef{r@#2}{\csletcs{r@#1}{r@#2}}{}%
}
\AtBeginDocument{%
  \konealiaslabel{M-fig:learning}{fig:learning}%
  \konealiaslabel{M-fig:perception_tools}{fig:perception_tools}%
  \konealiaslabel{M-fig:teaser}{fig:teaser}%
  \konealiaslabel{M-fig:toolusage}{fig:toolusage}%
  \konealiaslabel{M-tab:real_robot}{tab:real_robot}%
  \konealiaslabel{S-app:completion_protocol}{app:completion_protocol}%
  \konealiaslabel{S-app:gemini_rgb}{app:gemini_rgb}%
  \konealiaslabel{S-app:memory_ablation}{app:memory_ablation}%
  \konealiaslabel{S-app:protocol}{app:protocol}%
  \konealiaslabel{S-app:robocasa}{app:robocasa}%
  \konealiaslabel{S-app:robotwin}{app:robotwin}%
  \konealiaslabel{S-app:runtime_efficiency}{app:runtime_efficiency}%
  \konealiaslabel{S-app:toolfig}{app:toolfig}%
  \konealiaslabel{S-app:tools}{app:tools}%
  \konealiaslabel{S-app:training}{app:training}%
  \konealiaslabel{S-app:transfer}{app:transfer}%
  \konealiaslabel{S-tab:all}{tab:all}%
  \konealiaslabel{S-tab:category}{tab:category}%
  \konealiaslabel{S-tab:frontier_detail}{tab:frontier_detail}%
  \konealiaslabel{S-tab:student_interfaces}{tab:student_interfaces}%
}

%% file: sections/main.tex
\begingroup
\setlength{\intextsep}{8pt}
\begin{figure}[H]
\centering
\includegraphics[width=\linewidth]{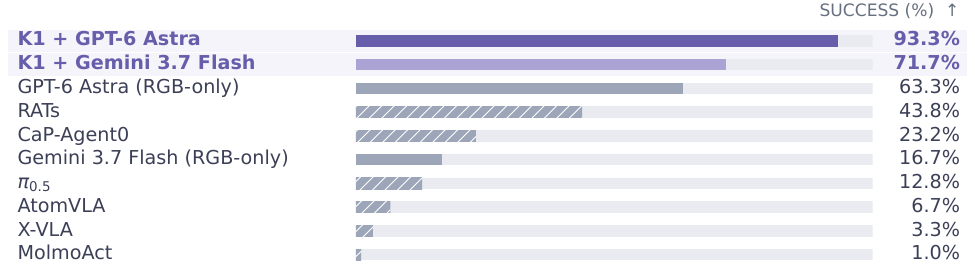}
\captionsetup{font=footnotesize,skip=5pt}
\caption{\textbf{Perception tools strengthen robot-use agents.} On 60 matched LIBERO-PRO configurations, K1 raises Gemini from 16.7\% to 71.7\% and Astra from 63.3\% to 93.3\%. Hatched bars show published references under their respective protocols~\cite{rats,atomvla}.}
\label{fig:teaser}
\end{figure}
\endgroup
\clearpage

\section{Introduction}
\begin{figure}[htbp]
\centering
\includegraphics[width=.975\textwidth]{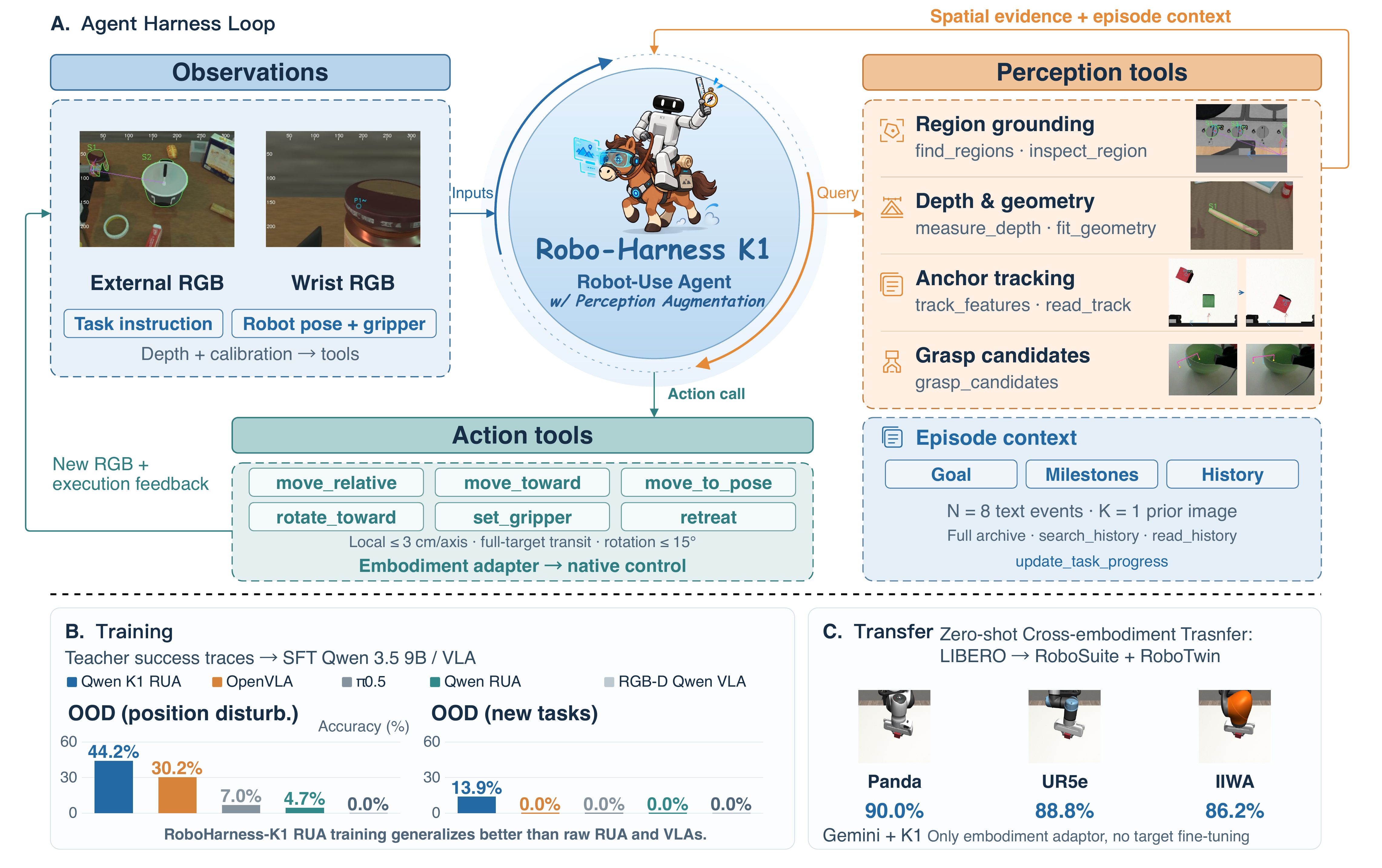}
\caption{\textbf{Robo-Harness K1.} (A) The agent queries perception tools and selects motions from RGB views, spatial evidence, and execution feedback. Depth and calibration support the tools, and adapters execute actions. (B) Epoch-5 new-state and held-out-task accuracy after training on 107 successful episodes. Methods share episodes but differ in inputs and interfaces. (C) Frozen Gemini with K1 on four shared RoboSuite tasks across three arm kinematics, all using the Panda gripper.}
\label{fig:architecture}
\end{figure}
Vision-language models (VLMs) recognize objects, follow instructions, and reason about visual relations~\cite{bai2025qwen3vl,rt2}. Manipulation demands more: the agent must tie a semantic goal to the metric geometry of objects and grippers.

Vision-language-action (VLA) models learn this link from action supervision~\cite{rt2,openvla,pi05}, at a cost. They depend on demonstration coverage, and adapting a VLM to motor outputs can degrade the representations it inherits~\cite{knowledge_insulation,blindvla}. LIBERO-PRO exposes the consequence: policies fail when object layouts or task conditions change~\cite{zhou2025liberopro}.

Robot-use agents (RUAs) take a different route. The VLM stays intact and controls the robot through an agentic loop of tool calls and observed effects~\cite{isola2026rua,via,showharness,inspectrobots}, and recent Astra results show its potential~\cite{robocurveastra}. With the same RGB-only interface, however, Gemini 3.7 Flash solves only 16.7\% of matched LIBERO-PRO cases, against 63.3\% for Astra (Figure~\ref{fig:teaser}). Gemini's failures are spatial rather than semantic. In one bowl-placement episode, it names the correct bowl and plate, then sweeps the gripper across 14\,cm in six consecutive calls while trying to center over the bowl. It reaches only 5.0\% on Spatial tasks, and in a third of its failed episodes it declares the task complete. The model understands the scene in 2D but cannot recover the 3D relations that motion requires. What separates a capable VLM from a working manipulation policy is accessible spatial evidence.

We propose \emph{perception as tools}. \method exposes four complementary sources of spatial evidence: grounded image regions, depth-backed geometry, persistent anchors, and grasp hypotheses. Measurements return as receipts and as overlays on the RGB views the VLM already consumes, so the model needs no new input modality. The agent decides what to inspect, selects a motion from the returned evidence, and measures again after acting. The harness supplies evidence and generic motions, not task-specific skills. A shared tool interface connects this loop to embodiment-specific sensing and control adapters (Figure~\ref{fig:architecture}).

The generality and effectiveness of the harness also make agentic post-training of a small RUA possible. Since evidence queries and physical actions share one language format, we find that the trace knowledge of a frontier model can be distilled into a small student. With 107 successful episodes, Qwen3.5-9B outperforms action-policy baselines trained on the same demonstrations, and mixed-environment training on traces from four environments raises its held-out accuracy from 11.1\% to 27.8\%.

Our contributions are threefold. (i) We introduce \method, a harness that connects VLM decisions to calibrated spatial measurements through perception tools. (ii) We show that K1 strengthens frozen frontier models and transfers to other arms, simulators, and a real robot. (iii) We show that agentic post-training can distill a frontier model's tool traces into a compact RUA. With 107 successful episodes, the student outperforms action-policy baselines and alone succeeds on held-out task conditions, and mixed-environment training improves it further.

\section{Related work}
\noindent\textbf{VLMs as robot policies.}
RT-2 transfers web-scale vision-language knowledge to action prediction~\cite{rt2}. OpenVLA provides an open action-token policy~\cite{openvla}, and $\pi_{0.5}$ adds heterogeneous co-training with a continuous action expert~\cite{pi05}. Action adaptation can erode pretrained capabilities~\cite{knowledge_insulation,blindvla}, which Actions as Language counters by aligning motor supervision with the language interface~\cite{actions_as_language}. Depth-aware VLMs instead add spatial inputs together with alignment training~\cite{sdvlm}. K1 keeps the VLM's interface unchanged: metric evidence arrives through tool calls, and the same serialized interface supervises the student.

\begin{figure}[!t]
\centering
\includegraphics[width=.975\textwidth]{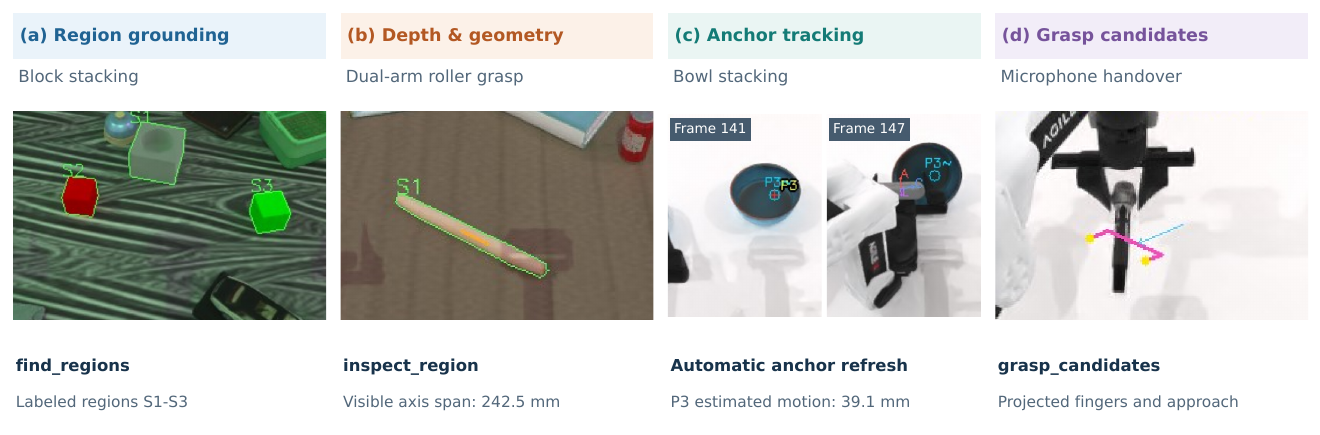}
\caption{\textbf{Four perceptual tool families on RoboTwin.} Head-camera crops show (a) block regions, (b) a roller's principal axis, (c) the same P3 anchor across frames, and (d) a microphone grasp hypothesis. Supplementary Section~\ref*{S-app:toolfig} details tool overlays.}
\label{fig:perception_tools}
\end{figure}

\noindent\textbf{Agentic manipulation.}
Language-model agents first controlled robots through intermediate abstractions: SayCan grounds plans in skill affordances~\cite{saycan}, Code as Policies generates executable programs~\cite{code_as_policies,cap_x,rats}, VoxPoser composes spatial value maps~\cite{voxposer}, and Inner Monologue feeds execution feedback into replanning~\cite{inner_monologue}. RUAs let the VLM command the robot directly: VIA, Show-Harness, and Inspect Robots expose robot commands through agentic interfaces~\cite{via,showharness,inspectrobots}, and GUAVA and Show-Harness distill such agents into compact policies~\cite{guava,showharness}. RoboHarness and Harness-VLA work at a different level, orchestrating heterogeneous policies through memory and planning~\cite{roboharnessmemory,harness_vla}. K1 builds on existing perception models, and its contribution lies in the interface. VoxPoser and code-as-policy systems call perception inside generated value maps or programs. K1 instead lets the VLM query evidence step by step, see the result in its own RGB views, and act directly. Prior RUA harnesses center on the action interface, whereas K1 makes perceptual queries part of the trace that the student learns.

\section{Method}
\label{sec:method}
\subsection{The agent harness loop}
Let $g$ be the instruction, $o_t$ the current RGB views and measured robot state, and $c_t$ the interaction context. At each decision, the VLM selects a tool $u$ and its arguments $a$:
\begin{equation}
(u,a)\sim\pi_\theta(\cdot\mid g,o_t,c_t).
\end{equation}
The tool set splits into information queries $\mathcal U_{\rm query}$ and physical commands $\mathcal U_{\rm act}$. Let $s_t$ denote the environment state. A query returns a receipt $r$ of spatial measurements, visual overlays, or retrieved context without moving the robot. Physical commands advance the robot and return an observation:
\begin{align}
u\in\mathcal U_{\rm query}:&\quad r=\mathcal H_{\rm query}(u,a;s_t),\\
u\in\mathcal U_{\rm act}:&\quad (r,o_{t+1})=\mathcal H_{\rm act}(u,a;s_t).
\end{align}
A context manager appends receipts and observations to the interaction history, so the agent can chain several queries to resolve uncertainty before committing to a motion.

The harness reads RGB-D, camera calibration, tool-center-point (TCP) pose, and gripper opening, while the VLM receives RGB views, robot state, and tool receipts. The VLM's input thus remains images and text, with metric measurements available on demand. Observation is partial: region measurements describe visible surfaces only, and tracking validity changes under motion or occlusion.

\subsection{Four perceptual tool families}
\label{sec:perception}

\noindent\textbf{Region grounding.}
Measurement starts with choosing what to measure. \tool{find\_regions} takes a text query and uses SAM~3~\cite{sam3} to return candidate masks, overlaid on the original image with persistent identifiers, and the agent picks among them visually. \tool{inspect\_region} refines a region within an agent-specified box. \tool{check\_region\_view} projects measured surface points into another available camera and checks depth consistency, providing cross-view verification.

\noindent\textbf{Depth and geometry.}
For a selected pixel $(u,v)$, \tool{measure\_depth} returns the calibrated world point:
\begin{equation}
x^w=T^w_c\left[D^c(u,v)(K^c)^{-1}(u,v,1)^\top;\,1\right],
\label{eq:unproject}
\end{equation}
where $D^c$ is depth, $K^c$ the camera intrinsics, and $T^w_c$ the camera-to-world transform. \tool{fit\_geometry} estimates planes or principal directions, and region inspection returns visible geometry. \tool{region\_relation} measures displacements between regions and from a region to the TCP. Point labels, principal axes, and TCP-to-surface lines tie each measurement to the RGB scene.

\noindent\textbf{Anchor tracking.}
Measurements should outlive the call that produced them. TAPNext++~\cite{tapnextpp} tracks image-point correspondences across frames. Current depth and camera pose lift them into world coordinates, separating camera motion from surface displacement. Each anchor keeps its identifier and a live validity flag. Uncertain correspondence invalidates the current measurement while the historical reference stays retrievable. The agent can refresh an anchor and read co-motion during a small lift as grasp evidence.

\noindent\textbf{Grasp candidates.}
A surface point does not specify a hand pose. Given a grounded region, GraspGen~\cite{graspgen} proposes parallel-jaw poses from its measured point cloud, and an embodiment adapter maps the provider's hand geometry to calibrated finger-pad coordinates. The agent receives candidate TCP positions, orientations, pregrasp waypoints, and projected schematics. It selects a candidate and executes the approach and closure itself through motion tools. Figure~\ref{fig:perception_tools} illustrates the four tool families on RoboTwin.

\subsection{Actions and memory}
\noindent\textbf{Adaptive motion granularity.}
\tool{move\_relative} applies a local correction of at most 3\,cm per axis. \tool{move\_to\_pose} and \tool{rotate\_toward} hold absolute targets across bounded steps. \tool{move\_toward} executes a longer target-directed transit that stops at tolerance, stall, or the execution budget. The agent thus alternates measured local alignment with larger transits. Each action reports achieved motion and residual error, and gripper feedback reports opening and settling. Supplementary Section~\ref*{S-app:tools} gives coordinates and tool contracts.

\noindent\textbf{Context and retrieval.}
The default context holds the instruction, current evidence, the last eight text exchanges, and the previous decision's images. Complete exchanges remain accessible through \tool{search\_history} and \tool{read\_history}. Each tool call carries a short decision note that records the evidence, intended motion, and expected outcome. The note makes tool selection inspectable and gives the student targets grounded in the information available at that decision.

\subsection{Agentic SFT of robot-use agents}
\label{sec:training}
Tool traces use the VLM's output format, so agentic post-training reduces to supervised fine-tuning (SFT) that distills a teacher's loop of evidence gathering and action into a student without a separate action head. Each teacher decision yields a context $c_j$ and one executable tool call serialized as $y_j$. Successful trajectories contain both information queries and physical commands. After filtering invalid or rejected target calls, we train the student with next-token supervision:
\begin{equation}
\mathcal L_{\rm agent}=-\sum_j\sum_{k\in y_j}\log\pi_\theta(y_{j,k}\mid c_j,y_{j,<k}).
\end{equation}
We adapt Qwen3.5-9B~\cite{qwen35} with language-side LoRA~\cite{lora} and a frozen vision encoder. At inference, the student calls the same harness tools as the teacher. The demonstration pool contains 107 successful Gemini episodes and excludes all held-out task conditions. Action-policy baselines draw their supervision from the same episodes but differ in inputs and execution interfaces (Supplementary Section~\ref*{S-app:training}).

We also continue the epoch-3 K1 adapter for three epochs on 316 successful trajectories from four environments and evaluate the student before and after (Section~\ref{sec:mixed_posttraining}).

\section{Experiments}
\label{sec:experiments}
We ask whether K1 strengthens frozen frontier agents and how they use its tools (Sections~\ref{sec:frontier} and~\ref{sec:perceptual_evidence}), whether it transfers across simulated embodiments and to a physical robot (Sections~\ref{sec:transfer} and~\ref{sec:real_robot}), whether agentic SFT distills tool traces into compact agents (Section~\ref{sec:student}), and which tools and how much history matter (Section~\ref{sec:tool_ablation}).

\subsection{Experimental setup}
\noindent\textbf{Benchmarks and metrics.}
LIBERO-PRO~\cite{zhou2025liberopro} is the primary benchmark for frontier deployment and student learning. Transfer experiments cover RoboSuite~\cite{robosuite}, RoboTwin~\cite{robotwin2}, RoboCasa~\cite{robocasa}, and a physical Piper arm. Simulation success follows each environment's native task criterion, and physical outcomes are confirmed by an operator and by image review.

\noindent\textbf{Frontier evaluation.}
We evaluate Gemini 3.7 Flash and GPT-6 Astra on Spatial, Object, and Goal tasks under position-swap and task perturbations. The full Gemini panel has 180 episodes: ten tasks, two perturbations, and three initial states per category. Its 60-case state-0 subset compares Gemini and Astra with the RGB-only Inspect Robots interface~\cite{inspectrobots} and with K1 on identical tasks and initial states. Both within-model comparisons evaluate complete interfaces, and Section~\ref{sec:tool_ablation} ablates individual tools. Published policies are references under their original protocols, not matched comparisons. K1 reads simulator depth and calibration. CaP-Agent0 and RATs also use depth, whereas the VLA references are RGB-only. The matched within-model results are our controlled evidence.

\noindent\textbf{Student evaluation.}
Of 139 successful teacher trajectories, 32 belong to twelve held-out conditions. The remaining 107, covering 43 conditions, train every primary method for five epochs. A condition includes its perturbation and instruction. Split A tests the 43 training configurations, B tests the same conditions at 43 new initial states and is the development set, and C tests the 12 held-out conditions at three states each (36 episodes). Baselines are OpenVLA-7B, official $\pi_{0.5}$, a Qwen3.5-9B continuous-action model (Qwen VLA), and an RGB-only text-action Qwen RUA. The complete Qwen VLA comparison uses its RGB-D variant. Untrained Qwen3.5-9B with K1 scores 0.0\% on all splits.

\noindent\textbf{Shared budgets and scoring.}
LIBERO-PRO episodes allow 1,200 native control steps at 20\,Hz and, for agents, at most 400 calls. Holds, gripper settling, and transits consume this control budget, while simulation pauses during observation and API waits. The paired Astra systems share tasks, initial states, camera placement, image resolution, and evaluator. Each completed policy outcome counts toward the success rate, and published references keep their trial counts.

\subsection{Frontier-model manipulation}
\label{sec:frontier}

\begin{table}[htbp]
\centering\small
\caption{\textbf{Full LIBERO-PRO evaluation (\%).} Each category averages position-swap and task perturbations. K1 uses 60 episodes per category (180 total), while published references use their original protocols~\cite{rats,atomvla}. Supplementary Table~\ref*{S-tab:frontier_detail} retains \mbox{the six-condition breakdown}.}
\label{tab:frontier}
\begin{tabular*}{\linewidth}{@{\extracolsep{\fill}}lrrrr@{}}
\toprule
Method & Object & Goal & Spatial & Avg.\\
\midrule
OpenVLA / $\pi_0$ & 0.0 & 0.0 & 0.0 & 0.0\\
$\pi_{0.5}$ & 9.0 & 19.0 & 10.5 & 12.8\\
MolmoAct & 3.0 & 0.0 & 0.0 & 1.0\\
NORA & 0.0 & 0.0 & 0.0 & 0.0\\
X-VLA & 5.0 & 5.0 & 0.0 & 3.3\\
AtomVLA & 5.0 & 6.5 & 8.5 & 6.7\\
CaP-Agent0 & 29.0 & 22.5 & 18.0 & 23.2\\
RATs & 62.0 & 39.5 & 30.0 & 43.8\\
\midrule
K1 + Gemini & \textbf{86.7} & \textbf{63.3} & \textbf{81.7} & \textbf{77.2}\\
\bottomrule
\end{tabular*}
\end{table}

\noindent\textbf{K1 strengthens both frontier models.}
On 60 matched configurations, K1 raises Gemini from 16.7\% to 71.7\% (+55.0 points) and Astra from 63.3\% to 93.3\% (+30.0 points, Figure~\ref{fig:teaser}). K1 solves 34 cases that RGB-only Gemini fails and twenty that RGB-only Astra fails, against one and two in reverse ($p=2.1\times10^{-9}$ and $p=1.2\times10^{-4}$, exact two-sided McNemar tests assuming independent pairs). Gemini's Spatial success rises from 5.0\% to 75.0\%, Object from 40.0\% to 80.0\%, and Goal from 5.0\% to 60.0\% (Supplementary Section~\ref*{S-app:gemini_rgb}). Astra also gains most on Spatial tasks, from 45.0\% to 90.0\%, where success hinges on recovering 3D relations from 2D views. The weaker model gains more, and Gemini with K1 beats RGB-only Astra by 8.3 points. The tools supply spatial ability that weaker models lack.

\noindent\textbf{Full Gemini evaluation.}
With three initial states per task condition (180 episodes), Gemini with K1 succeeds in 139 (77.2\%): 86.7\% on Object, 81.7\% on Spatial, and 63.3\% on Goal (Table~\ref{tab:frontier}). The 71.7\% above is the state-0 subset of this evaluation. Object and Spatial tasks hinge on target localization and relative geometry, which the tools measure directly. Goal tasks add articulated and multi-stage contact, whose difficulty extends beyond spatial perception. Supplementary Sections~\ref*{S-app:protocol} and~\ref*{S-app:completion_protocol} give per-condition results and the post-release scoring diagnostic.

\subsection{Understanding K1's behavior}
\begin{figure}[htbp]
\centering
\includegraphics[width=.68\linewidth,trim=0 4 0 0,clip]{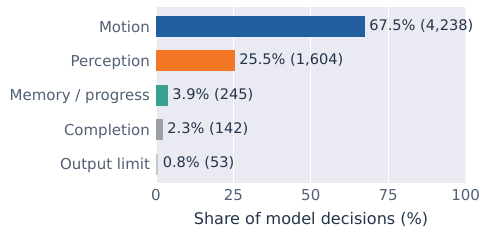}
\caption{\textbf{How the agent allocates decisions.} Counts and shares of 6,282 Gemini responses over 180 LIBERO-PRO episodes, including failures. Perception and memory together account for \mbox{29.4\% of decisions}.}
\label{fig:toolusage}
\end{figure}
\noindent\textbf{Perception is used throughout execution.}
\label{sec:perceptual_evidence}
Figure~\ref{fig:toolusage} summarizes 180 Gemini LIBERO-PRO trajectories. Motion tools account for 67.5\% of 6,282 decisions, perception for 25.5\%, and memory/progress for 3.9\%. Perception is not a one-time scene scan: of the 1,604 perception requests, 1,036 (64.6\%) occur after the initial simulation frame, in 159 of 180 trajectories. The most frequent queries measure depth (733), ground or inspect regions (571), and propose grasps (274). Grounding fixes what to measure, depth supplies metric targets, and grasp proposals guide contact. The explicit tracking tools are never requested. Persistent visual anchors are instead refreshed automatically between calls.

\noindent\textbf{Evidence-guided recovery.}
In a paired bowl-placement case, Astra with K1 grounds the target bowl, inspects it from the wrist camera, closes the gripper, and verifies the grasp with a 25\,mm lift. It then measures the bowl-to-plate relation and refines the placement through local corrections. K1 completes the task and passes the post-release check, whereas RGB-only Astra ends with an incorrect placement.

\noindent\textbf{When measurement is not enough.}
A failed plate-placement case marks the current limit. Late corrections achieve only 26\% and 10\% of the requested motion, a 6.7\,cm alignment error persists near timeout, and the agent keeps issuing local corrections until the budget runs out. The evidence exposes the error, but the agent does not change its strategy. Such cases call for policies that turn measured errors into revised approaches and grasps.

\subsection{Transfer across environments and embodiments}
\label{sec:transfer}

\begin{table}[htbp]
\centering\small
\caption{\textbf{RoboSuite success (\%).} K1: twenty trials per task/arm. Cube row: mean of lift, restack, and stack. CaP-Agent0 and RATs are published references~\cite{rats}. UR5e/IIWA use the Panda gripper.}
\label{tab:transfer}
\begin{tabular*}{\linewidth}{@{\extracolsep{\fill}}lrrrrr@{}}
\toprule
& \multicolumn{2}{c}{References} & \multicolumn{3}{c}{K1 + Gemini}\\
\cmidrule(lr){2-3}\cmidrule(l){4-6}
Task & CaP & RATs & Panda & UR5e & IIWA\\
\midrule
Cubes (3-task avg.) & 49.3 & 63.3 & \textbf{100} & \textbf{100} & \textbf{100}\\
Nut assembly & 0.0 & 0.0 & \textbf{60.0} & 55.0 & 45.0\\
Wiping & \textbf{100} & \textbf{100} & 65.0 & --- & ---\\
Two-arm handover & \textbf{24.0} & 20.0 & 5.0 & --- & ---\\
Two-arm lifting & 10.0 & 34.0 & \textbf{75.0} & --- & ---\\
\midrule
Four-task avg. & 37.0 & 47.5 & \textbf{90.0} & 88.8 & 86.2\\
All seven tasks & 40.3 & 49.1 & \textbf{72.1} & --- & ---\\
\bottomrule
\end{tabular*}
\end{table}

\begin{figure}[htbp]
\centering
\includegraphics[width=.64\linewidth,trim=0 4 0 0,clip]{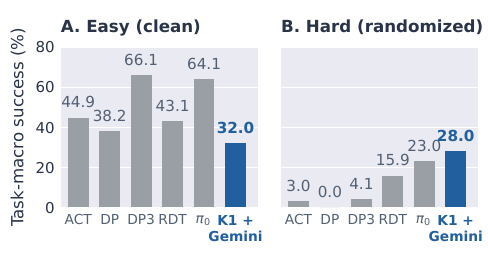}
\caption{\textbf{Bimanual transfer on RoboTwin.} Task-macro success under Easy (left) and Hard (right) conditions. Gray bars are trained-policy references~\cite{robotwin2}, and blue bars use frozen Gemini with K1. Supplementary material gives task coverage and evaluation counts.}
\label{fig:robotwin_main}
\end{figure}
\noindent\textbf{RoboSuite: new arm kinematics.}
We deploy frozen Gemini with K1 and change only the sensing and control adapters. The VLM, perception providers, spatial conventions, and tool interface carry over unchanged. On four shared tasks, success is 90.0\% on Panda, 88.8\% on UR5e, and 86.2\% on IIWA (Table~\ref{tab:transfer}). UR5e and IIWA keep the Panda gripper, which isolates the change in arm kinematics. All three arms solve every cube task, and nut assembly accounts for the remaining spread. On seven Panda tasks, K1 reaches 72.1\%, including 65.0\% on wiping and 75.0\% on two-arm lifting. Handover reaches only 5.0\%: the agent struggles to coordinate release with a secure receiving grasp. Once perception is supplied, the bottleneck shifts to contact-rich assembly and two-arm coordination. Supplementary Section~\ref*{S-app:transfer} gives the full table and adapters, and Section~\ref*{S-app:robocasa} reports RoboCasa results.

\noindent\textbf{RoboTwin: bimanual transfer.}
Ten AgileX dual-arm tasks in SAPIEN add another simulator and embodiment. Zero-shot K1 reaches task-macro success of 32.0\% on Easy and 28.0\% on Hard, a four-point gap (Figure~\ref{fig:robotwin_main}). Hard randomizes lighting, textures, clutter, and table height~\cite{robotwin2}. Under the same shift, trained-policy references fall from 43.1\% to 15.9\% for RDT and from 64.1\% to 23.0\% for $\pi_0$. K1 thus trails these references on Easy but is competitive on Hard. Plausibly, the tools let the agent measure geometry again after appearance and execution change. Per-task results and reference protocols appear in Supplementary Section~\ref*{S-app:robotwin}.

The ten tasks were selected after preliminary testing. Of 200 planned episodes, 18 fail at initialization and are excluded, leaving 96 Easy and 86 Hard scored evaluations.

\subsection{Real-robot experiments}
\label{sec:real_robot}
\noindent\textbf{Two short-horizon tasks and one long-horizon task.}
The short tasks place a can into a bowl and a bowl onto a plate, each needing one grasp-and-place sequence. The long task, two-object sorting, places the blue block in the yellow container and the red block in the green bowl.

\noindent\textbf{Platform and comparison.}
We use a Piper arm with a parallel-jaw gripper and two external RGB-D cameras. Both methods run Gemini 3.7 Flash on the same Piper control and inverse-kinematics adapters. The Clean baseline follows uni-robot-agent~\cite{uniagent} and receives RGB and measured arm pose through four tools: scene capture, motion, hold, and finish. K1 adds perception tools and interaction memory.

\begin{figure}[htbp]
\centering
\includegraphics[width=\textwidth]{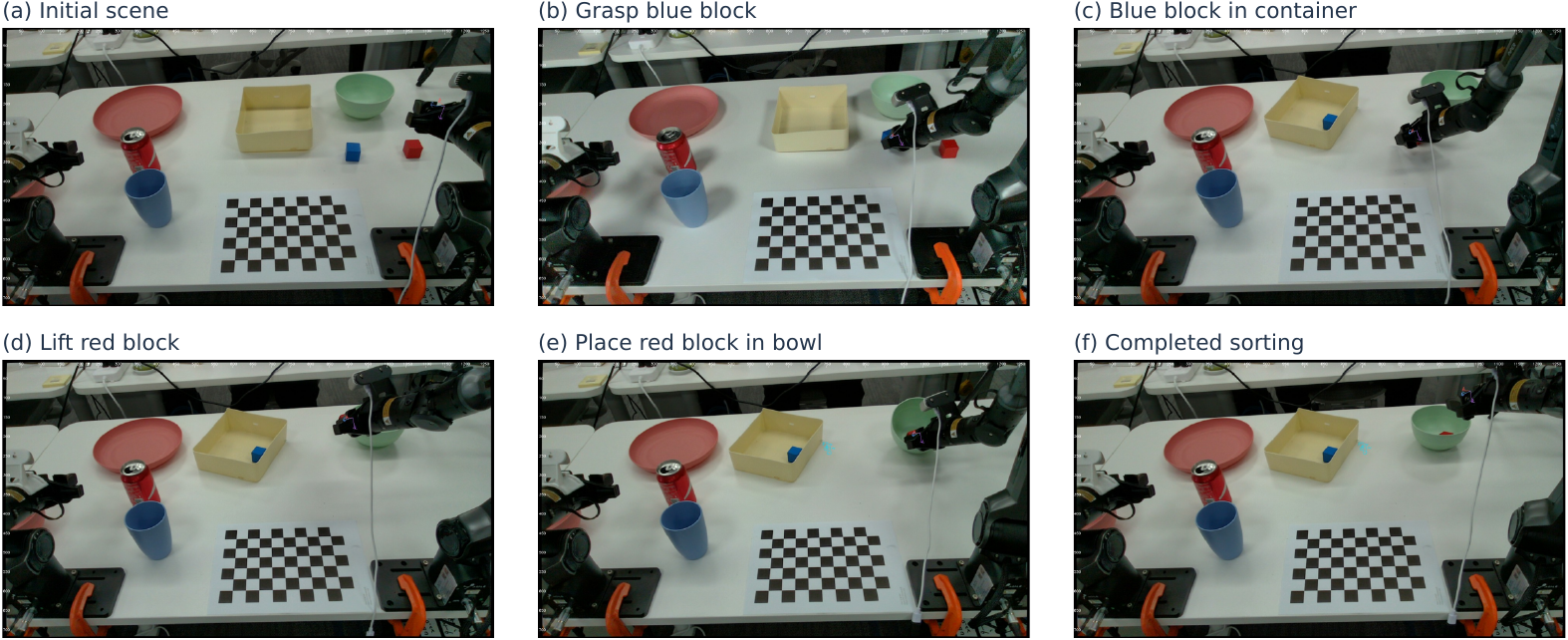}
\caption{\textbf{Real-robot experiments: long-horizon two-object sorting on a Piper arm.} Head-camera observations show placement of the blue block in the yellow container, followed by the red block in the green bowl. The snapshots span one recorded K1 trial.}
\label{fig:real_robot_sorting}
\end{figure}

\begin{table}[t]
\centering\small
\setlength{\tabcolsep}{3pt}
\caption{\textbf{Short- and long-horizon Piper tasks.} Three concluded trials per method/task. Success gives completed trials, and progress is the mean manually assessed score (0 to 2).}
\label{tab:real_robot}
\begin{tabular*}{\linewidth}{@{\extracolsep{\fill}}lrrrr@{}}
\toprule
& \multicolumn{2}{c}{Success} & \multicolumn{2}{c}{Progress}\\
\cmidrule(lr){2-3}\cmidrule(l){4-5}
Task & Clean & K1 & Clean & K1\\
\midrule
\multicolumn{5}{@{}l}{\emph{Short horizon}}\\
Can into bowl & 1/3 & \textbf{2/3} & 1.00 & \textbf{1.67}\\
Bowl onto plate & 0/3 & \textbf{3/3} & 0.00 & \textbf{2.00}\\
\multicolumn{5}{@{}l}{\emph{Long horizon}}\\
Two-object sorting & 0/3 & \textbf{3/3} & 0.50 & \textbf{2.00}\\
\midrule
Overall & 1/9 & \textbf{8/9} & --- & ---\\
Success rate (\%) & 11.1 & \textbf{88.9} & --- & ---\\
\bottomrule
\end{tabular*}
\end{table}

\noindent\textbf{Short-horizon manipulation.}
K1 completes five of six short-task trials, whereas Clean completes one (Table~\ref{tab:real_robot}). Can placement improves from 1/3 to 2/3, and bowl placement from 0/3 to 3/3. Progress scores, which rate approach, grasp-and-lift, transport, and stable placement on a 0 to 2 scale, agree: 1.67 on can placement and 2.00 on bowl placement for K1, against 1.00 and 0.00 for Clean.

\noindent\textbf{Long-horizon sorting.}
K1 completes both placements in all three sorting trials (Figure~\ref{fig:real_robot_sorting}), whereas Clean completes none. Mean progress is 2.00 for K1 and 0.50 for Clean. Overall, K1 completes eight of nine trials and Clean one.

\noindent\textbf{Execution protocol.}
Both methods have a 50-call budget for single-object trials and a 100-call budget for sorting. The archived K1 sorting trial finishes at 88 calls. One bowl trial includes external completion review and gripper-unit guidance, and an operator-interrupted can attempt is recorded separately. This study therefore demonstrates feasibility and leaves fully autonomous evaluation to future work.

\subsection{Agentic SFT of compact RUAs}
\label{sec:student}

\begin{figure}[htbp]
\centering
\includegraphics[width=\textwidth]{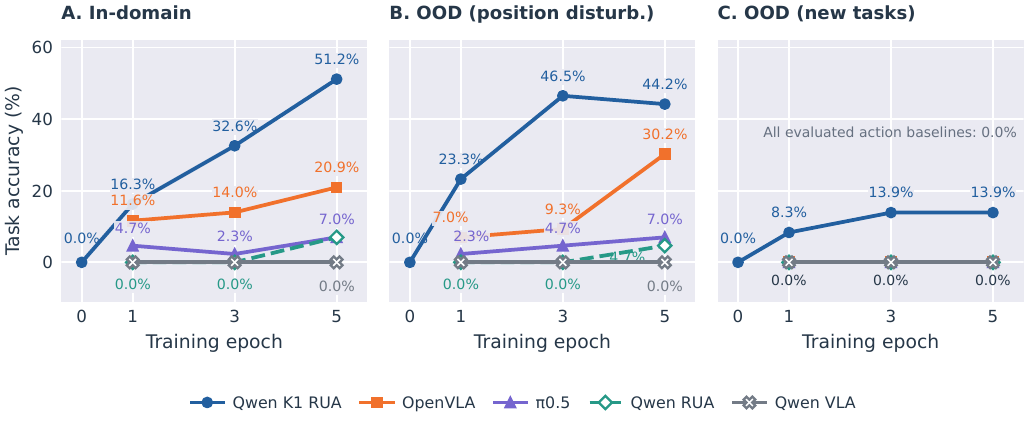}
\caption{\textbf{Learning from 107 successful episodes.} A: exact training configurations. B: new initial states, used as the development set. C: held-out task conditions. Qwen K1 RUA learns tool interactions, and Qwen RUA learns RGB-only text actions. The Qwen VLA epoch-5 comparison uses the fully evaluated RGB-D variant. Supplementary Tables~\ref*{S-tab:all} and~\ref*{S-tab:category} provide checkpoint counts and category results.}
\label{fig:learning}
\end{figure}

\begin{table}[htbp]
\centering\small
\caption{\textbf{New-state generalization (B, \%).} Epoch 5, with 15 Spatial, 16 Object, and 12 Goal cases. All methods share \mbox{107 source demonstrations}.}
\label{tab:student_category_main}
\begin{tabular*}{\linewidth}{@{\extracolsep{\fill}}lrrrr@{}}
\toprule
Student & Spatial & Object & Goal & Avg.\\
\midrule
Qwen K1 RUA & \textbf{46.7} & 43.8 & \textbf{41.7} & \textbf{44.2}\\
Qwen RUA & 0.0 & 6.2 & 8.3 & 4.7\\
Qwen VLA (RGB-D) & 0.0 & 0.0 & 0.0 & 0.0\\
OpenVLA & 6.7 & \textbf{62.5} & 16.7 & 30.2\\
$\pi_{0.5}$ & 6.7 & 6.2 & 8.3 & 7.0\\
\bottomrule
\end{tabular*}
\end{table}

\noindent\textbf{What each student learns.}
The 107 demonstrations yield 2,482 K1 tool-call targets, 1,670 RGB-only RUA action targets, and 40,808 native action frames (10,241 chunk-prediction windows). Both RUAs train rank-16 language LoRA with frozen vision encoders: the K1 student observes RGB, state, and tool evidence, whereas the text-action RUA observes RGB alone. Qwen VLA uses RGB-D and proprioception, while OpenVLA and $\pi_{0.5}$ keep their pretrained action representations (Supplementary Table~\ref*{S-tab:student_interfaces}). Inputs preserve pre-action observations and earlier attempts within successful trajectories, so recovery examples are retained. VLA action normalization uses only the 107 training episodes.

\noindent\textbf{Agentic SFT outperforms action supervision.}
After five epochs, Qwen K1 RUA reaches 51.2\% on A and 44.2\% on B, against 20.9\% and 30.2\% for OpenVLA and 7.0\% and 7.0\% for $\pi_{0.5}$ (Figure~\ref{fig:learning}). The RGB-D Qwen VLA scores 0.0\% on all three splits. The text-action format alone does not explain the gain: the RGB-only Qwen RUA reaches 7.0\% on A and 4.7\% on B, and invalid actions terminate 35.2\% of its evaluations. The value lies in the complete perceptual interface, not merely in expressing actions as text.

\noindent\textbf{Generalization to new states.}
On split B (Table~\ref{tab:student_category_main}), K1 leads OpenVLA by 40.0 points on Spatial and 25.0 on Goal, while OpenVLA leads on Object (62.5\% versus 43.8\%). Agentic SFT thus pays off most in reusable spatial decisions rather than uniformly across task categories. A likely reason is that tool traces teach a measure-then-move procedure that carries over to new layouts, whereas action supervision fits scene-specific trajectories.

\noindent\textbf{Generalization to held-out tasks.}
K1 is the only evaluated method that succeeds on split C, reaching 13.9\% (5 of 36 episodes), while every baseline scores 0.0\%. Its successes span four held-out conditions and all three categories: three Spatial, one Object, and one Goal. None of these conditions appears in the training demonstrations.

\noindent\textbf{Learning dynamics.}
K1's success on training configurations climbs from 16.3\% at epoch 1 to 32.6\% at epoch 3 and 51.2\% at epoch 5. Generalization saturates earlier: new-state success peaks at 46.5\% at epoch 3, and held-out-task success stays at 13.9\% from epoch 3 to 5. Later optimization mainly fits the training configurations (Figure~\ref{fig:learning}).

\begin{table}[htbp]
\centering\small
\caption{\textbf{Mixed-environment continuation (\%).} Fresh evaluation of the epoch-3 checkpoint and its continued adapter. Supplementary material gives exact counts, including the \mbox{39/40-episode RoboTwin panels}.}
\label{tab:mixed_main}
\begin{tabular*}{\linewidth}{@{\extracolsep{\fill}}lrr@{}}
\toprule
Evaluation & Before & After\\
\midrule
LIBERO-PRO B: new states & 42.9 & \textbf{52.4}\\
LIBERO-PRO C: held-out tasks & 11.1 & \textbf{27.8}\\
RoboSuite & 28.3 & \textbf{50.0}\\
RoboCasa & 8.3 & 4.2\\
RoboTwin & 2.6 & 2.5\\
\bottomrule
\end{tabular*}
\end{table}

\noindent\textbf{Mixed-environment SFT.}
\label{sec:mixed_posttraining}
We continue SFT of the epoch-3 K1 student for three epochs on 316 successful trajectories: 107 from LIBERO-PRO, 142 from RoboSuite, 32 from RoboCasa, and 35 from RoboTwin. Fresh before/after evaluations show gains on LIBERO-PRO B and C and on RoboSuite, while RoboCasa and RoboTwin stay below 10\% (Table~\ref{tab:mixed_main}). Sample and environment diversity thus improves generalization on tasks whose skills the student already has, here single-arm tabletop manipulation. It does not by itself unlock harder skills: mobile and bimanual control, each with fewer than 40 trajectories, remain unsolved. Separating optimization time from data diversity would require a matched source-only continuation.

\subsection{Ablations and analysis}
\label{sec:tool_ablation}
\begin{table}[!htb]
\centering\small
\caption{\textbf{Perception-tool ablations (\%).} Thirty matched episodes per environment. Region tools also return geometry, and their removal disables dependent grasp tools.}
\label{tab:tool_family_main}
\begin{tabular*}{\linewidth}{@{\extracolsep{\fill}}lrr@{}}
\toprule
Configuration & LIBERO-PRO & RoboTwin\\
\midrule
Full perception tools & \textbf{70.0} & \textbf{30.0}\\
Remove depth/local geometry & 56.7 & 16.7\\
Remove regions and dependents & 50.0 & 13.3\\
Remove cross-frame tracking & 40.0 & 20.0\\
Remove grasp candidates & 60.0 & 13.3\\
\bottomrule
\end{tabular*}

\end{table}

\noindent\textbf{Perception-tool contributions.}
We remove one tool family at a time on matched 30-episode Gemini panels in LIBERO-PRO and RoboTwin. Each removal updates the tool instructions, so the agent sees only the remaining tools.

The full tool set performs best in both environments (Table~\ref{tab:tool_family_main}). On LIBERO-PRO, removing cross-frame tracking hurts most, from 70.0\% to 40.0\%. RoboTwin is most sensitive to losing region tools or grasp candidates, dropping from 30.0\% to 13.3\% in either case. This ablation removes two things: the explicit tracking tools, which the agent never requests (Section~\ref{sec:perceptual_evidence}), and the persistent visual anchors that the harness keeps current across frames. The drop comes from the latter, because measurements then expire after each motion. Bimanual manipulation leans more on grounded regions and candidate hand poses. With 30 episodes per cell, small differences are within noise.

\begin{figure}[!htb]
\centering
\includegraphics[width=.76\linewidth]{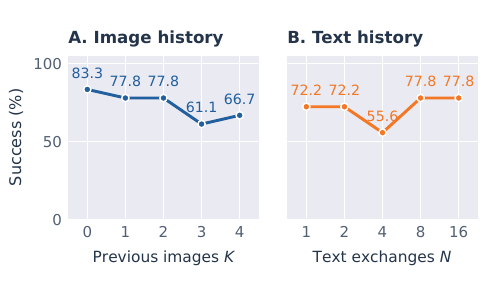}
\caption{\textbf{Memory-context ablation.} Success on 18 matched Gemini cases per setting. Left: previous image rounds $K$ at $N=8$ text exchanges. Right: $N$ at $K=1$, on a log$_2$ axis. Retrieval remains available throughout.}
\label{fig:memory_ablation}
\end{figure}

\noindent\textbf{Memory-context selection.}
On 18 matched cases (Figure~\ref{fig:memory_ablation}), the default context of one previous image round and eight text exchanges reaches 77.8\% success in 33.0 calls. Text windows of one or two exchanges reach 72.2\% but need 70.0 and 72.6 calls. The text-history curve is nonmonotonic, and a single outcome shifts accuracy by 5.6 points, so the clearest benefit of a moderate history is fewer repeated interactions. Supplementary Section~\ref*{S-app:memory_ablation} reports all settings.

More image history does not help. Extending it from one to three rounds lowers success to 61.1\% and raises mean calls to 47.4, while current images alone reach 83.3\% in 28.3 calls. Current visual evidence matters most.

\noindent\textbf{Runtime cost.}
On the 60 paired Astra cases, K1 takes 7.31 minutes and 45.9 calls per episode, against 4.47 minutes and 24.8 calls for RGB-only control. Gemini with K1 averages 5.16 minutes over its 180 cases. Model requests consume 83\% and 77\% of episode time for Astra and Gemini with K1, respectively, so serving latency and decision count are the main efficiency targets (Supplementary Section~\ref*{S-app:runtime_efficiency}).

\FloatBarrier
\Needspace{10\baselineskip}
\section{Conclusion}
\method exposes perception as tools so that a VLM can recover the 3D relations that motion requires. Without weight updates, K1 raises Gemini from 16.7\% to 71.7\% and Astra from 63.3\% to 93.3\% on LIBERO-PRO, and it transfers to new embodiments, simulators, and a real robot. Agentic post-training distills its tool traces into a compact agent that outperforms action-policy baselines given the same demonstrations. K1 still relies on calibrated depth and visible geometry, and fine contact, coordinated release, and held-out accuracy leave clear room for improvement.

%% file: sections/supplement.tex
\section{Method and Tool Details}
\label{app:tools}

\subsection{Tool families and contracts}
Table~\ref{tab:tools} summarizes the public tool families. Image measurements specify their camera, frame, and pixel-coordinate convention. Segmentation and tracking attach visual evidence to user-labeled inspection regions.\looseness=-1

\begin{table*}[!t]
\centering\small
\caption[Perception and action tool contracts]{Perception, control, and memory tools in the evaluated harness. Task completion is assessed separately using the \mbox{protocol in Section~\ref{app:completion_protocol}}.}\label{tab:tools}
\begin{tabularx}{\textwidth}{@{}>{\raggedright\arraybackslash}p{.32\linewidth}>{\raggedright\arraybackslash}p{\dimexpr.68\linewidth-2\tabcolsep\relax}@{}}
\toprule
Tool & Returned evidence or executed command\\
\midrule
\tool{find\_regions} & Text-prompted SAM 3 masks from a current original RGB image, with RGB-D summaries and stable S identifiers. Multiple candidates require model-side disambiguation.\\
\tool{inspect\_region} & Box-prompted segmentation and region refresh. Returns the visible surface geometry within the supplied box.\\
\tool{measure\_depth} & Back-projects a valid visible pixel to world coordinates and returns TCP-relative displacement and a trackable surface-point reference.\\
\tool{fit\_geometry} & Fits a plane or unsigned principal direction to visible geometry in a selected image region.\\
\tool{region\_relation} & Measures region-to-robot or region-to-region displacement and available plane relations.\\
\tool{check\_region\_view} & Projects a measured region into the other available camera and checks depth consistency.\\
\tool{track\_features}, \tool{read\_track} & Starts and queries image correspondences with visibility estimates. Point and region measurement tools also maintain anchors.\\
\tool{inspect\_crop} & Enlarges a selected region of an existing RGB observation.\\
\tool{grasp\_candidates} & Returns nominal pad-aligned TCP poses, quaternions, pregrasp points, and projected schematics for agent selection.\\
\tool{move\_relative} & Executes a supplied relative translation in world or gripper coordinates. Each requested component is bounded by $\delta_{\rm axis}$. Optional rotation is a bounded world rotation vector.\\
\tool{move\_to\_pose} & Creates or resumes an absolute TCP pose target. Executes one bounded translation/rotation chunk and preserves the gripper command. Explicit new coordinates supersede an old target identifier.\\
\tool{rotate\_toward}, \tool{align\_axis} & Moves toward an absolute quaternion or aligns a calibrated gripper axis to a requested direction. Rotation is at most $15^\circ$ per call, with positional anchoring.\\
\tool{move\_toward} & Sends an absolute target to native control within one call, stopping on tolerance, stall, or execution budget. Collision avoidance remains part of motion selection.\\
\tool{set\_gripper} & Uses 0 for closed and 1 for open while holding pose; returns settling status and measured jaw gap.\\
\tool{retreat} & Takes a bounded step toward an earlier observed pose; the agent assesses current path clearance.\\
\tool{align\_region\_references} & Computes a translation hypothesis aligning source/destination visible references, with a model-chosen offset and axes. Relies on current measurement and a stable carry relation.\\
\tool{search\_history}, \tool{read\_history} & Searches completed episode calls and retrieves paginated evidence, optionally including archived images. Historical frames remain explicitly labeled.\\
\tool{update\_task\_progress}, \tool{remember} & Stores model-authored milestones, uncertainties, and working notes alongside the original task.\\
\tool{done} & Submits a completion request under the configured completion protocol (Section~\ref{app:completion_protocol}); remaining budgets are unchanged.\\
\bottomrule
\end{tabularx}
\end{table*}

\subsection{Segmentation, tracking, and geometry}
Text search returns a small set of nonredundant candidates with persistent identifiers. Region measurements combine the current RGB-D observation with camera calibration to produce visible surface geometry in robot coordinates. Anchors are refreshed across views and time; lost or depth-inconsistent correspondence prompts reinspection. This keeps object identity, pixel location, and metric position distinct throughout execution.

\subsection{Grasp proposal and ranking}
The grasp provider receives the observed region cloud and maps candidate hand poses into the calibrated robot frame. It aligns the nominal contact reference with the finger-pad midpoint, uses parallel-jaw symmetry to reduce unnecessary rotation, and removes near-duplicate hypotheses. Candidate approach paths are checked against the available depth views. The harness returns up to three geometric hypotheses with visual schematics, while the model selects and verifies the grasp through subsequent observations.

\subsection{Control and feedback}
Local tools execute bounded translation or rotation, while persistent targets preserve the intended pose across repeated calls. Full-target transit stops on arrival or stalled progress. Every motion reports the achieved pose and residual so the model can continue, revise the target, or choose a new approach. The model receives execution evidence rather than internal controller diagnostics or private grasp scores.

\subsection{Perception-tool illustrations}
\label{app:toolfig}
\begin{figure*}[!t]
\centering
\includegraphics[width=.95\linewidth]{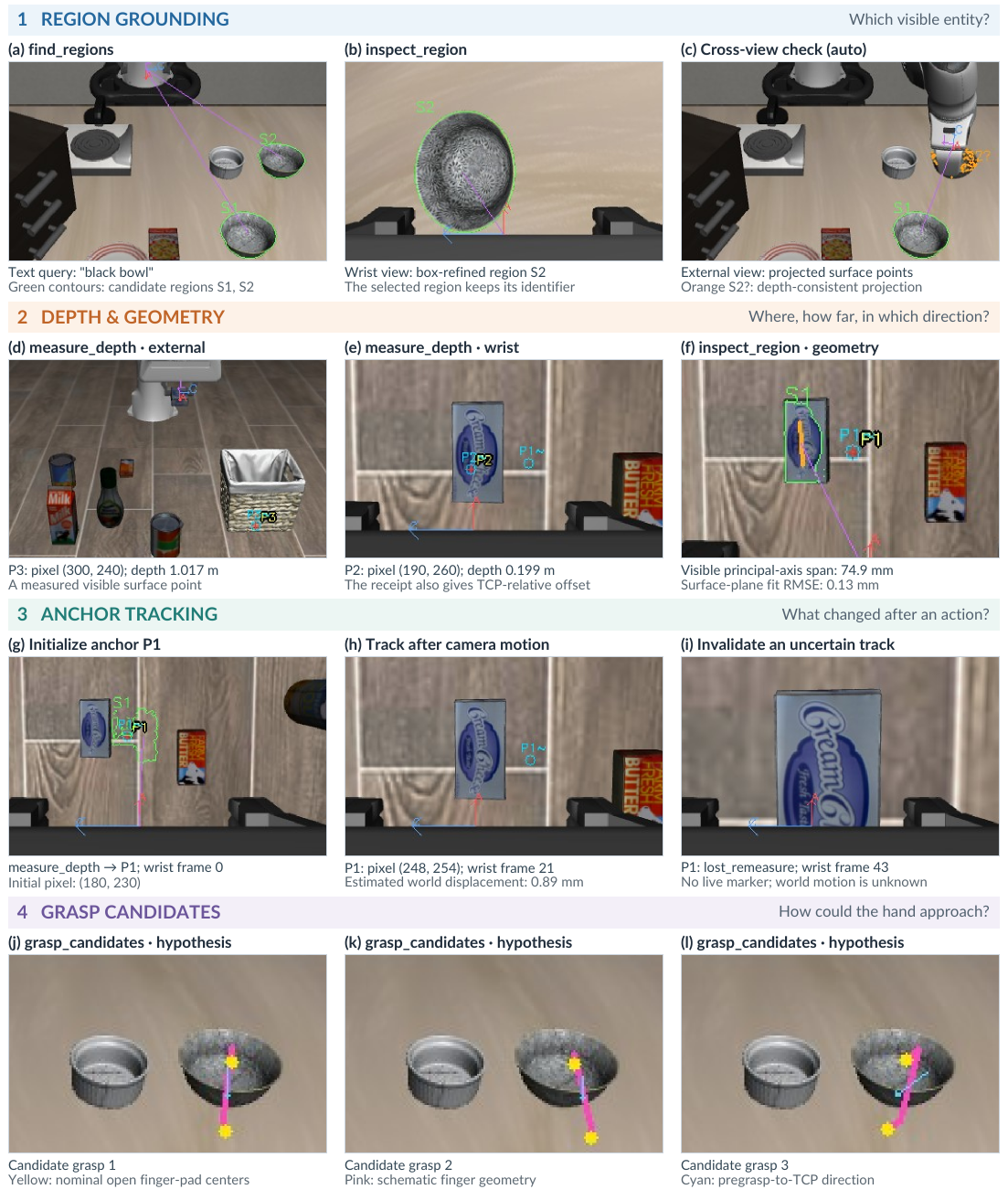}
\caption[Perception-tool visual atlas]{\textbf{Perception as tools, illustrated by recorded model-visible feedback.} Each row corresponds to one tool family. (a to c) Text grounding, box refinement, and depth-consistent projection into the other existing view. (d to f) Metric points and visible-surface geometry, with selected receipt values below the images. (g to i) The same wrist-camera anchor changes pixel location but has only 0.89\,mm estimated world displacement, before becoming invalid. Anchor coordinates update automatically between calls. (j to l) Three grasp hypotheses from one tool response, shown as enlarged crops of proposed hand poses. Green contours denote regions, orange lines denote principal axes, and purple lines denote TCP-to-surface references.}
\label{fig:perception_tools_atlas}
\end{figure*}

Figure~\ref{fig:perception_tools_atlas} provides a complementary LIBERO tool atlas. Main-paper Figure~\ref*{M-fig:perception_tools} illustrates the same four tool families on RoboTwin: block stacking, roller grasping, bowl stacking, and microphone handover. Its tracking example follows P3 during bowl manipulation, whereas this atlas follows P1 under wrist-camera motion. All panels retain the recorded RGB inputs and measurement coordinates.\looseness=-1

\noindent\textbf{Visual encoding.}
Green S contours denote segmentation estimates. Orange principal-axis lines describe visible geometry. Purple lines connect TCP and surface references. The gripper's A/C/L axes denote approach, closing, and lateral directions. P labels mark measured or tracked surface points, with a tilde denoting tracking estimates. In panels g to i, the same anchor P1 changes pixel location after the wrist moves while maintaining sub-millimeter estimated world displacement, until the tracker reports the anchor as lost.

Panel f reports the region's visible principal-axis span and surface-plane residual from the same \tool{inspect\_region} call. Panels c and g to i illustrate automatic cross-view and anchor maintenance. The grasp panels show nominal open finger-pad centers (yellow), simplified finger lines (pink), and pregrasp-to-TCP directions (cyan). The displayed hypotheses come from one response.

\FloatBarrier
\section{Training and Baselines}
\label{app:training}

\subsection{Data construction and filtering}
The training pool contains successful teacher trajectories. Of 139 successful Gemini episodes, 32 belong to withheld conditions, leaving 107 episodes from 43 conditions. The native-action dataset contains 40,808 control frames. Eight-step windows sampled at stride four give 10,241 training windows for chunk-prediction models, which execute four actions before observing again. OpenVLA trains on every native frame and executes one action at a time.

The tool dataset uses original pre-action requests, tool schemas, and model-visible images. It excludes twelve execution-error targets, sixteen responses with missing or multiple calls, and 54 rejected completions, retaining 2,482 targets (96.8\% of 2,564 decisions). Targets are executable calls with their short decision notes, and inputs come from the model-visible interaction record. Decision-note delimiters are ordinary text tokens.

Inputs preserve the teacher's observed history, including earlier rejected completion requests. Depth and action supervision are regenerated by replaying the native actions.

\subsection{Architectures and optimization}
\begin{table*}[t]
\centering\small
\caption{\textbf{Student inputs and supervision.} Every method uses the same 107 source episodes and five training epochs. Inputs, target granularity, and control interfaces follow the respective model designs. Detailed optimization settings are in Section~\ref{app:training}.}
\label{tab:student_interfaces}
\begin{tabularx}{\textwidth}{@{}lYY@{}}
\toprule
Student & Model-visible input & Supervision and execution\\
\midrule
Qwen K1 RUA & RGB views, robot state, tool evidence and history & Tool calls with short decision notes; queries and generic motions\\
Qwen RUA & Current external RGB and instruction & JSON physical commands; local step, target transit, gripper\\
Qwen VLA (RGB-D) & RGB, measured proprioception, local depth features & Eight-step continuous native-action chunks\\
OpenVLA-7B & Current external RGB and instruction & Discretized native-action tokens; one action per observation\\
$\pi_{0.5}$ & External/wrist RGB and discretized robot state & Flow-matching action chunks; four actions before replanning\\
\bottomrule
\end{tabularx}
\end{table*}
\begin{table*}[!t]
\centering\small
\caption[Post-training configurations]{Adaptation recipe. All runs use seed 17, batch size 1 with accumulation 8, and save checkpoints at epochs 1, 3, and 5.}
\label{tab:trainconfig}
\begin{tabularx}{\linewidth}{@{}>{\raggedright\arraybackslash}p{.21\linewidth}Y>{\raggedright\arraybackslash}p{.17\linewidth}>{\raggedleft\arraybackslash}p{.17\linewidth}@{}}
\toprule
Run & Trainable components & Learning rate & Examples/epoch\\
\midrule
Qwen K1 RUA & Language LoRA $r=16$, $\alpha=32$, dropout .05 & $10^{-4}$ & 2,482\\
Qwen RUA & Same language LoRA, native text output head & $10^{-4}$ & 1,670\\
Qwen VLA (RGB-only / RGB-D) & Same LoRA plus numeric-state/action head, optional depth encoder & $10^{-4}$ & 10,241\\
$\pi_{0.5}$ & Language LoRA $r=16$, $\alpha=32$, plus full action expert/projections & $2.5\times10^{-5}$ & 10,241\\
OpenVLA & All-linear LoRA $r=32$, $\alpha=16$, dropout .05 & $5\times10^{-4}$ & 40,808\\
\bottomrule
\end{tabularx}
\end{table*}

\noindent\textbf{Qwen K1 RUA.}
The base is Qwen3.5-9B with a frozen vision encoder and a rank-16 language adapter. Cross-entropy is applied to assistant tool-call targets. At evaluation time, the student receives the same harness observations and tool schemas as the teacher.

\noindent\textbf{Qwen VLA.}
The Qwen VLA shares the Qwen3.5-9B backbone and predicts eight-step native-action chunks from the language state, proprioception, and optional depth features. It provides a controlled comparison between tool-call supervision and direct continuous-action regression on the same backbone.

\noindent\textbf{Original OpenVLA.}
OpenVLA-7B receives the current external RGB image and task instruction. Its tokenizer predicts discretized native actions, with normalization statistics fitted on the 107 training trajectories. One action executes before the next observation.

\noindent\textbf{$\pi_{0.5}$.}
The $\pi_{0.5}$ baseline starts from the official general checkpoint and retains its flow-matching action expert. It receives external and wrist RGB with robot state, predicts action chunks, and executes four actions before replanning. Normalization statistics come from the same training pool.

Training took approximately 26.6 hours for Qwen K1 RUA, 5.7 hours for $\pi_{0.5}$, and 19.1 hours for OpenVLA.

\subsection{RGB-to-action baseline}
\label{app:rgbaction}
The RGB text-action baseline starts from the Qwen3.5-9B base checkpoint. Its input consists of the task instruction and one current unannotated external RGB image.

We retain all 1,670 teacher decisions that produce physical execution: local motion, target-transit, and gripper commands. Labels encode the commanded action at each decision boundary. The output is a JSON action with fields for mode, translation, rotation, and gripper state. Three modes are supported: \tool{step} (bounded local motion), \tool{target} (full-target transit), and \tool{gripper} (open/close). Robot state is available internally to the executor.

Training uses the same language LoRA settings as the K1 student, with a frozen vision encoder and cross-entropy loss on assistant action tokens only. Five epochs produce 1,045 optimizer updates. Inference is greedy with no constrained decoding. Invalid actions count as failures.\looseness=-1

Epochs 1, 3, and 5 are evaluated on the same manifest of 43 A, 43 B, and 36 C cases. At epoch 5, the 122 evaluations end in 74 budget terminations, 43 invalid actions (35.2\%), and five native successes. Format errors thus account for a substantial part of this baseline's failures. The complete counts appear in Table~\ref{tab:all}.\looseness=-1

\subsection{RGB baseline and paired interface comparison}
\label{app:rgb}
We reuse the Inspect Robots agent and tool implementation~\citep{inspectrobots}.
The LIBERO adapter supplies external and wrist RGB, measured end-effector position/orientation, and normalized gripper opening, without depth measurements.

Its tools are \tool{move\_to}, \tool{done}, and \tool{give\_up}. The agent specifies world XYZ and optional roll/pitch/yaw relative to the reset orientation. The adapter converts this to the native rotational command using
\begin{equation}
R_{\rm target}=R_z(\mathrm{yaw})R_y(\mathrm{pitch})R_x(\mathrm{roll})R_{\rm reset}.
\end{equation}
The tool interpolates the requested pose into smaller targets. The adapter holds the final target for twelve frames.\looseness=-1

The paired systems share task, initial state, camera placement, image resolution, native horizon, maximum calls, and evaluator. Table~\ref{tab:interfaces} lists the differences in perception, memory, motion, and completion feedback. The comparison evaluates K1 as a complete interface. Gemini and Astra use the same RGB-only implementation and settings, changing only the model.

\begin{table*}[!t]
\centering\small
\caption[Interfaces in paired frontier comparisons]{Interface differences in the within-model comparisons for Gemini and Astra.}
\label{tab:interfaces}
\begin{tabularx}{\linewidth}{@{}>{\raggedright\arraybackslash}p{.24\linewidth}YY@{}}
\toprule
Property & Original RGB interface & K1\\
\midrule
Scene information & Two RGB views & Same two views plus depth-backed measurements\\
Text history & Original complete history & Last eight exchanges plus retrieval\\
Image history & Current and previous decision & Current and previous decision, optional retrieved images\\
Motion & Original interpolated pose tool & Local corrections, persistent poses, full-target transit\\
Perception & Model interprets RGB & Explicit region, depth, tracking, grasp tools\\
Progress memory & Dialogue & Dialogue plus model-maintained milestones\\
Completion & Ends on completion submission & Checks completion and permits correction within the remaining budget\\
\bottomrule
\end{tabularx}
\end{table*}

\section{LIBERO-PRO Evaluation Details}
\label{app:protocol}

\subsection{Benchmark protocol and data splits}
LIBERO evaluation and student training use a fixed harness configuration. The frontier evaluation contains 180 Gemini K1 episodes and three matched 60-episode comparisons: Gemini RGB-only and both Astra interfaces. Student evaluation covers 122 cases, including twelve task conditions held out from fine-tuning.

The full evaluation comprises ten tasks in each of Spatial/Object/Goal, two variants (swap and task), and three initial states, giving 180 trials and 30 per reported column. The paired frontier comparison covers all ten task indices at state 0 in all six suites, giving 60 trials per system and twenty per category. Gemini's K1 result is taken from the full evaluation. Gemini RGB-only and both Astra interfaces use the same configurations. Published LIBERO-PRO rows from RATs Table 1 use 100 trials per column and 600 overall and serve as reference results.\looseness=-1

The MolmoAct, NORA, X-VLA, and AtomVLA rows in Table~\ref{tab:frontier_detail} come from \citet{atomvla}, Table~IV. Their averages use the six displayed Position and Task perturbations for Object, Goal, and Spatial. Published rows retain their source protocols.

The Gemini full evaluation yields 139/180 (77.2\%) native successes, with category totals of 52/60 (86.7\%) Object, 38/60 (63.3\%) Goal, and 49/60 (81.7\%) Spatial. The paired subset yields 10/60 for Gemini RGB, 38/60 for Astra RGB, 43/60 for Gemini K1, and 56/60 for Astra K1.

The LIBERO-PRO benchmark contains 50 stored initial states per task. We select initial states by their stored state-file indices. The teacher evaluation uses 0, 1, and 2. Student B uses 3. Student C uses 40, 41, and 42.

\noindent\textbf{Student split inventory.}
Table~\ref{tab:split} lists all fine-tuning-covered and held-out conditions by suite. A task condition is identified jointly by suite variant and task index. A and B each contain 15 Spatial, 16 Object, and 12 Goal conditions. C contains four per category, evaluated at three states each.

\begin{table}[!t]
\centering\small
\caption[Training and held-out task inventory]{Task-index inventory. A uses one actually trained initial state for each listed condition. B uses the same condition at state 3. C uses states 40, 41, and 42.}
\label{tab:split}
\begin{tabularx}{\linewidth}{@{}lYl@{}}
\toprule
Suite & Trained (A/B) & Held out (C)\\
\midrule
goal\_swap & 0, 1, 4, 7--9 & 2, 6\\
goal\_task & 3--5, 7--9 & 2, 6\\
object\_swap & 2--9 & 0, 1\\
object\_task & 2--9 & 0, 1\\
spatial\_swap & 1, 2, 5--9 & 0, 3\\
spatial\_task & 1--4, 6--9 & 0, 5\\
\bottomrule
\end{tabularx}
\end{table}

A uses a successful training demonstration for each condition, prioritizing state 0: 33 cases use state 0, nine use state 1, and one uses state 2. Five other conditions have no successful teacher demonstration and are excluded from A/B/C: Goal swap 3 and 5, Goal task 0 and 1, and Spatial swap 4.\looseness=-1

C tests new task conditions, including new object relations within shared scenes and manipulation skills. Both swap and task perturbations are represented in fine-tuning.

\subsection{Execution and scoring}
\label{app:completion_protocol}
\noindent\textbf{Completion feedback.}
K1 uses the simulator's Boolean task-completion check when the model submits \tool{done}. An unsuccessful request returns control to the model within the remaining budget, while native success terminates the episode. Both RGB-only baselines terminate on \tool{done} or native success. Section~\ref{app:po_ablation} measures the contribution of perception and completion feedback.

Every local case has 1,200 native control steps at 20\,Hz. Agentic cases additionally have a 400-call limit. All controller holds, gripper settling, and transit steps consume this budget. Physics pauses during observation, tool computation, and API waiting.

Native simulator success is the primary metric, and every completed policy outcome is retained. The additional Astra and Gemini RGB-only episodes use terminal native success.

As a stability check, we release the gripper after replay and test whether success persists. This diagnostic gives 131/180 (72.8\%) for Gemini K1, compared with the primary native-success result of 139/180 (77.2\%). All headline results use the primary metric reported in the main paper.

\subsection{Full LIBERO-PRO comparison}
\begin{table*}[!t]
\centering\small
\setlength{\tabcolsep}{5pt}
\caption[LIBERO-PRO task accuracy]{\textbf{LIBERO-PRO task accuracy (\%).} Position denotes the position-swap suite. MolmoAct, NORA, X-VLA, and AtomVLA are from \citet{atomvla}, Table~IV, and other VLA and CaP rows are from RATs~\citep{rats}. Averages use the six displayed conditions.}
\label{tab:frontier_detail}
\begin{tabularx}{\linewidth}{l*{6}{R}r}
\toprule
 & \multicolumn{2}{c}{Object} & \multicolumn{2}{c}{Goal} & \multicolumn{2}{c}{Spatial} & Average\\
Method & Position & Task & Position & Task & Position & Task & \\
\midrule
\multicolumn{8}{l}{\emph{VLA Methods}}\\
OpenVLA & 0.0 & 0.0 & 0.0 & 0.0 & 0.0 & 0.0 & 0.0\\
$\pi_0$ & 0.0 & 0.0 & 0.0 & 0.0 & 0.0 & 0.0 & 0.0\\
$\pi_{0.5}$ & 17.0 & 1.0 & 38.0 & 0.0 & 20.0 & 1.0 & 12.8\\
MolmoAct & 6.0 & 0.0 & 0.0 & 0.0 & 0.0 & 0.0 & 1.0\\
NORA & 0.0 & 0.0 & 0.0 & 0.0 & 0.0 & 0.0 & 0.0\\
X-VLA & 2.0 & 8.0 & 1.0 & 9.0 & 0.0 & 0.0 & 3.3\\
AtomVLA & 10.0 & 0.0 & 2.0 & 11.0 & 16.0 & 1.0 & 6.7\\
\midrule
\multicolumn{8}{l}{\emph{Code as Policy Methods}}\\
CaP-Agent0 & 27.0 & 31.0 & 29.0 & 16.0 & 13.0 & 23.0 & 23.2\\
RATs & 61.0 & 63.0 & 43.0 & 36.0 & 29.0 & 31.0 & 43.8\\
\midrule
\multicolumn{8}{l}{\emph{RUA Methods}}\\
\rowcolor{kpale}\method + Gemini & \textbf{80.0} & \textbf{93.3} & \textbf{70.0} & \textbf{56.7} & \textbf{73.3} & \textbf{90.0} & \textbf{77.2}\\
\bottomrule
\end{tabularx}
\end{table*}

\subsection{Paired frontier comparison}
\label{app:frontier_summary}
Table~\ref{tab:paired} gives the category breakdown for the four local systems summarized in main-paper Figure~\ref*{M-fig:teaser}. The systems use the same 60 configurations of task, variant, and initial state.

\noindent\textbf{A task-complete, category-balanced subset.}
The matched evaluation covers all ten task indices in each category, with both swap and task variants at initial state 0. Spatial, Object, and Goal each contribute twenty cases, and swap and task perturbations each contribute thirty. This deterministic subset gives the six strata equal weight, matching the full Gemini evaluation's category balance.

\noindent\textbf{Relation to the full evaluation.}
Gemini with K1 achieves 139/180 (77.2\%) across three initial states and 43/60 (71.7\%) on the matched state-0 subset. Main-paper Figure~\ref*{M-fig:teaser} uses the latter for comparison with Gemini RGB-only and both Astra interfaces, while Table~\ref{tab:frontier_detail} retains all 180 cases.\looseness=-1

\noindent\textbf{Comparison scope.}
The four local systems share task identities and initial states, giving a controlled comparison of the complete K1 and RGB-only interfaces. Published results provide broader benchmark context under their original protocols. K1 and the code-as-policy systems use depth, while the VLA references use RGB. Paired Gemini and Astra outcomes and their significance tests are reported in Section~\ref{app:allresults}.

\begin{table}[!t]
\centering\small
\caption[Paired system comparison on matched cases]{\textbf{Paired system comparison: task accuracy (\%).} Tasks and initial states are matched. Each system is evaluated on twenty cases per category, sixty in total. Each within-model comparison changes perception, memory, control interface, and \mbox{completion feedback together}.}
\label{tab:paired}
\begin{tabularx}{\linewidth}{l*{4}{R}}
\toprule
System & Spatial & Object & Goal & Avg.\\
\midrule
Gemini (RGB) & 5.0 & 40.0 & 5.0 & 16.7\\
Astra (RGB) & 45.0 & 80.0 & 65.0 & 63.3\\
Gemini (K1) & 75.0 & 80.0 & 60.0 & 71.7\\
\rowcolor{kpale}Astra (K1) & \textbf{90.0} & \textbf{100.0} & \textbf{90.0} & \textbf{93.3}\\
\bottomrule
\end{tabularx}
\end{table}

\subsection{Full checkpoint results}
\label{app:allresults}
\begin{table*}[!t]
\centering\small
\caption[Full checkpoint evaluation results]{Complete checkpoint results used in the main-paper comparison.}\label{tab:all}
\begin{tabularx}{\textwidth}{@{}>{\raggedright\arraybackslash}p{\dimexpr.31\linewidth-2\tabcolsep\relax}>{\raggedright\arraybackslash}p{\dimexpr.09\linewidth-2\tabcolsep\relax}*{3}{>{\raggedleft\arraybackslash}p{\dimexpr.20\linewidth-1.333333\tabcolsep\relax}}@{}}
\toprule
Model & Epoch & A & B & C\\
\midrule

Qwen K1 RUA (base) & 0 & 0/43 (0.0\%) & 0/43 (0.0\%) & 0/36 (0.0\%)\\
Qwen K1 RUA & 1 & 7/43 (16.3\%) & 10/43 (23.3\%) & 3/36 (8.3\%)\\
Qwen K1 RUA & 3 & 14/43 (32.6\%) & \textbf{20/43 (46.5\%)} & \textbf{5/36 (13.9\%)}\\
Qwen K1 RUA & 5 & \textbf{22/43 (51.2\%)} & 19/43 (44.2\%) & \textbf{5/36 (13.9\%)}\\
Qwen RUA & 1 & 0/43 (0.0\%) & 0/43 (0.0\%) & 0/36 (0.0\%)\\
Qwen RUA & 3 & 0/43 (0.0\%) & 0/43 (0.0\%) & 0/36 (0.0\%)\\
Qwen RUA & 5 & 3/43 (7.0\%) & 2/43 (4.7\%) & 0/36 (0.0\%)\\
OpenVLA-7B & 1 & 5/43 (11.6\%) & 3/43 (7.0\%) & 0/36 (0.0\%)\\
OpenVLA-7B & 3 & 6/43 (14.0\%) & 4/43 (9.3\%) & 0/36 (0.0\%)\\
OpenVLA-7B & 5 & 9/43 (20.9\%) & 13/43 (30.2\%) & 0/36 (0.0\%)\\
$\pi_{0.5}$ & 1 & 2/43 (4.7\%) & 1/43 (2.3\%) & 0/36 (0.0\%)\\
$\pi_{0.5}$ & 3 & 1/43 (2.3\%) & 2/43 (4.7\%) & 0/36 (0.0\%)\\
$\pi_{0.5}$ & 5 & 3/43 (7.0\%) & 3/43 (7.0\%) & 0/36 (0.0\%)\\
Qwen VLA (RGB-D) & 5 & 0/43 (0.0\%) & 0/43 (0.0\%) & 0/36 (0.0\%)\\
\bottomrule
\end{tabularx}
\end{table*}
Qwen K1 RUA, Qwen RUA, OpenVLA, and $\pi_{0.5}$ have complete epoch 1/3/5 evaluations. Qwen VLA (RGB-D) is reported at epoch 5, where it completes all 122 cases without a success.

\begin{table*}[!t]
\centering\small
\caption[Epoch-5 accuracy by task category]{Epoch-5 success by task category. Denominators are 15/16/12 for A and B and 12 per category for C.}\label{tab:category}
\begin{tabularx}{\textwidth}{@{}>{\raggedright\arraybackslash}p{\dimexpr.31\linewidth-2\tabcolsep\relax}>{\raggedright\arraybackslash}p{\dimexpr.09\linewidth-2\tabcolsep\relax}*{3}{>{\raggedleft\arraybackslash}p{\dimexpr.20\linewidth-1.333333\tabcolsep\relax}}@{}}
\toprule
Model & Group & Spatial & Object & Goal\\
\midrule

Qwen K1 RUA & A & \textbf{6/15 (40.0\%)} & \textbf{11/16 (68.8\%)} & \textbf{5/12 (41.7\%)}\\
Qwen K1 RUA & B & \textbf{7/15 (46.7\%)} & 7/16 (43.8\%) & \textbf{5/12 (41.7\%)}\\
Qwen K1 RUA & C & \textbf{3/12 (25.0\%)} & \textbf{1/12 (8.3\%)} & \textbf{1/12 (8.3\%)}\\
Qwen RUA & A & 0/15 (0.0\%) & 1/16 (6.2\%) & 2/12 (16.7\%)\\
Qwen RUA & B & 0/15 (0.0\%) & 1/16 (6.2\%) & 1/12 (8.3\%)\\
Qwen RUA & C & 0/12 (0.0\%) & 0/12 (0.0\%) & 0/12 (0.0\%)\\
OpenVLA-7B & A & 0/15 (0.0\%) & 7/16 (43.8\%) & 2/12 (16.7\%)\\
OpenVLA-7B & B & 1/15 (6.7\%) & \textbf{10/16 (62.5\%)} & 2/12 (16.7\%)\\
OpenVLA-7B & C & 0/12 (0.0\%) & 0/12 (0.0\%) & 0/12 (0.0\%)\\
$\pi_{0.5}$ & A & 0/15 (0.0\%) & 2/16 (12.5\%) & 1/12 (8.3\%)\\
$\pi_{0.5}$ & B & 1/15 (6.7\%) & 1/16 (6.2\%) & 1/12 (8.3\%)\\
$\pi_{0.5}$ & C & 0/12 (0.0\%) & 0/12 (0.0\%) & 0/12 (0.0\%)\\
\bottomrule
\end{tabularx}
\end{table*}
The epoch-5 agent's five C successes cover four task conditions: Goal task 6/state 40, Object swap 0/state 42, Spatial swap 0/states 40 and 41, and Spatial task 5/state 42.

For Gemini, K1 succeeds on 34 cases where RGB-only control fails, while the reverse occurs on one. Both succeed on nine and both fail on sixteen. For Astra, K1 succeeds on twenty cases where RGB-only control fails, while the reverse occurs on two. Both succeed on 36 and both fail on two. The exact two-sided McNemar test gives $2\sum_{i=0}^{1}\binom{35}{i}2^{-35}\approx2.1\times10^{-9}$ for Gemini and $2\sum_{i=0}^{2}\binom{22}{i}2^{-22}\approx1.2\times10^{-4}$ for Astra, indicating significant improvements under the independent-pair assumption.

\section{Transfer Experiments}
\label{app:transfer}

\subsection{RoboSuite: cross-embodiment transfer}
\begin{table*}[!t]
\centering\small
\setlength{\tabcolsep}{5pt}
\caption[RoboSuite zero-shot transfer accuracy]{\textbf{RoboSuite zero-shot transfer accuracy (\%).} K1 uses Gemini without target fine-tuning. CaP-Agent0 and RATs are published results~\citep{rats}, and -- denotes untested tasks.}
\label{tab:transfer_detail}
\begin{tabular*}{\linewidth}{@{\extracolsep{\fill}}lrrrrr@{}}
\toprule
 & \multicolumn{2}{c}{References~\citep{rats}} & \multicolumn{3}{c}{K1 + Gemini (ours)}\\
Task/Methods or Arms & CaP-Agent0 & + RATs skills & Panda & UR5e & IIWA\\
\midrule
Cube lifting & 68.0 & 84.0 & \textbf{100.0} & \textbf{100.0} & \textbf{100.0}\\
Cube restacking & 34.0 & 46.0 & \textbf{100.0} & \textbf{100.0} & \textbf{100.0}\\
Cube stacking & 46.0 & 60.0 & \textbf{100.0} & \textbf{100.0} & \textbf{100.0}\\
Nut assembly & 0.0 & 0.0 & \textbf{60.0} & 55.0 & 45.0\\
Spill wiping & \textbf{100.0} & \textbf{100.0} & 65.0 & -- & --\\
Two-arm handover & \textbf{24.0} & 20.0 & 5.0 & -- & --\\
Two-arm lifting & 10.0 & 34.0 & \textbf{75.0} & -- & --\\
\midrule
Common four-task avg. & 37.0 & 47.5 & \textbf{90.0} & 88.8 & 86.2\\
Seven-task avg. & 40.3 & 49.1 & \textbf{72.1} & -- & --\\
\bottomrule
\end{tabular*}
\end{table*}

\noindent\textbf{Setup.}
The RoboSuite adapter retains the K1 runtime and frozen Gemini 3.7 Flash model. Only the sensing and control adapters change, without target-task training.\looseness=-1

We reuse RATs/CaP task construction and native success checks for cube lifting, restacking, stacking, square-nut assembly, spill wiping, two-arm handover, and two-arm lifting. Panda covers all seven, while UR5e and IIWA cover the four shared single-arm tasks. The three arms use the Panda gripper, isolating arm kinematics in the cross-embodiment comparison. Wiping uses the task's attached sponge.

The adapter supplies calibrated scene and wrist RGB-D views and maps generic Cartesian commands to each robot's operational-space controller. Dual-arm commands advance both effectors together and return a residual for each arm. Tools that do not apply to a task, such as gripper commands during wiping, are hidden from the model.

\noindent\textbf{Trials and budgets.}
Each tested task/embodiment combination has 20 trials over five reset configurations. K1 uses at most 400 model decisions. Native horizons are 1,500 steps for cube and nut tasks, 4,000 for wiping, and 5,000 for dual-arm tasks. Table~\ref{tab:transfercounts} reports every evaluated combination.

\noindent\textbf{VLA baseline mappings.}
OpenVLA and $\pi_{0.5}$ use their official base checkpoints and action normalization. OpenVLA receives scene RGB and produces Cartesian actions. $\pi_{0.5}$ receives scene and wrist RGB with robot state and produces joint-space action chunks. The available normalization supports Panda and UR5e, so $\pi_{0.5}$ is not evaluated on IIWA. These rows measure the official base models; Figure~\ref*{M-fig:learning} evaluates the LIBERO-trained variants.

\noindent\textbf{Results.}
The common four-task mean uses cube lifting, restacking, stacking, and nut assembly. K1 reaches 72/80 on Panda, 71/80 on UR5e, and 69/80 on IIWA. Across all seven Panda tasks, it reaches 101/140. The task-level counts in Table~\ref{tab:transfercounts} show that nut assembly and handover account for most remaining errors.

\begin{table*}[!t]
\centering\small
\caption[Complete RoboSuite transfer results]{Complete transfer results. Each tested task has five reset configurations and four rollouts per configuration. The VLA rows use official base checkpoints. Their grouped rows have zero successes on every constituent task, with 20 trials per task. The four shared tasks are cube lifting, restacking, stacking, and nut assembly.}\label{tab:transfercounts}
\begin{tabularx}{\textwidth}{@{}>{\raggedright\arraybackslash}p{\dimexpr.20\linewidth-2\tabcolsep\relax}>{\raggedright\arraybackslash}p{\dimexpr.12\linewidth-2\tabcolsep\relax}>{\raggedright\arraybackslash}p{\dimexpr.28\linewidth-2\tabcolsep\relax}*{2}{>{\raggedleft\arraybackslash}p{\dimexpr.20\linewidth-\tabcolsep\relax}}@{}}
\toprule Method & Robot & Task & Successes / trials & Accuracy (\%)\\\midrule
K1 + Gemini & Panda & Cube lifting & 20/20 & 100.0\\
K1 + Gemini & Panda & Cube restacking & 20/20 & 100.0\\
K1 + Gemini & Panda & Cube stacking & 20/20 & 100.0\\
K1 + Gemini & Panda & Nut assembly & 12/20 & 60.0\\
K1 + Gemini & Panda & Spill wiping & 13/20 & 65.0\\
K1 + Gemini & Panda & Two-arm handover & 1/20 & 5.0\\
K1 + Gemini & Panda & Two-arm lifting & 15/20 & 75.0\\
K1 + Gemini & Panda & Tested-task total & 101/140 & 72.1\\\midrule
K1 + Gemini & UR5e & Cube lifting & 20/20 & 100.0\\
K1 + Gemini & UR5e & Cube restacking & 20/20 & 100.0\\
K1 + Gemini & UR5e & Cube stacking & 20/20 & 100.0\\
K1 + Gemini & UR5e & Nut assembly & 11/20 & 55.0\\
K1 + Gemini & UR5e & Tested-task total & 71/80 & 88.8\\\midrule
K1 + Gemini & IIWA & Cube lifting & 20/20 & 100.0\\
K1 + Gemini & IIWA & Cube restacking & 20/20 & 100.0\\
K1 + Gemini & IIWA & Cube stacking & 20/20 & 100.0\\
K1 + Gemini & IIWA & Nut assembly & 9/20 & 45.0\\
K1 + Gemini & IIWA & Tested-task total & 69/80 & 86.2\\\midrule
OpenVLA base & Panda & Four shared tasks and spill wiping & 0/100 & 0.0\\
OpenVLA base & UR5e & Four shared tasks & 0/80 & 0.0\\
OpenVLA base & IIWA & Four shared tasks & 0/80 & 0.0\\
$\pi_{0.5}$ base & Panda & Four shared tasks and spill wiping & 0/100 & 0.0\\
$\pi_{0.5}$ base & UR5e & Four shared tasks & 0/80 & 0.0\\
$\pi_{0.5}$ base & IIWA & Unevaluated & -- & --\\
\bottomrule
\end{tabularx}
\end{table*}

\subsection{RoboTwin: bimanual transfer}
\label{app:robotwin}

\noindent\textbf{Setting and transfer scope.}
RoboTwin uses Aloha AgileX dual arms in SAPIEN. Easy denotes \texttt{demo\_clean}, and Hard denotes \texttt{demo\_randomized}, which varies clutter, illumination, textures, and table height~\citep{robotwin2}. K1 supplies native head and two wrist RGB-D views, robot state, perceptual tools, and generic single- or dual-arm commands. Zero-shot denotes deployment without target-task model updates. The interface adapter was tested on RoboTwin tasks.

\noindent\textbf{Task and episode selection.}
We evaluate ten tasks from the RoboTwin benchmark, with ten planned episodes per task under both Easy and Hard conditions. Eighteen episodes fail native initialization before policy execution, leaving 96 Easy and 86 Hard episodes. Every initialized episode is scored.

\noindent\textbf{Execution and aggregation.}
The model-call limit is 400, with text history $N=8$, image history $K=1$, and a 3\,cm per-axis local translation limit. We report the mean task accuracy used in the main paper: 32.0\% on Easy and 28.0\% on Hard. Episode-level totals are 31/96 and 25/86, respectively.

\begin{table*}[!t]
\centering\small
\setlength{\tabcolsep}{5pt}
\caption[K1 + Gemini per-task results on RoboTwin]{\textbf{K1 + Gemini on RoboTwin.} Scores are successes/retained episodes followed by accuracy. Excluded counts are initialization failures in Easy/Hard order.}
\label{tab:robotwin_tasks}
\begin{tabular*}{\linewidth}{@{\extracolsep{\fill}}lrrcr@{}}
\toprule
Task & Easy & Hard & Excluded & Action limit\\
\midrule
Grab Roller & 8/10 (80.0\%) & 9/10 (90.0\%) & 0/0 & 400\\
Handover Mic & 2/10 (20.0\%) & 0/10 (0.0\%) & 0/0 & 600\\
Lift Pot & 2/10 (20.0\%) & 2/10 (20.0\%) & 0/0 & 400\\
Move Can Pot & 4/8 (50.0\%) & 5/8 (62.5\%) & 2/2 & 400\\
Place Dual Shoes & 0/8 (0.0\%) & 0/6 (0.0\%) & 2/4 & 600\\
Place Phone Stand & 2/10 (20.0\%) & 0/8 (0.0\%) & 0/2 & 400\\
Put Bottles Dustbin & 0/10 (0.0\%) & 0/8 (0.0\%) & 0/2 & 1700\\
Put Object Cabinet & 4/10 (40.0\%) & 0/8 (0.0\%) & 0/2 & 700\\
Stack Blocks Two & 6/10 (60.0\%) & 7/8 (87.5\%) & 0/2 & 800\\
Stack Bowls Two & 3/10 (30.0\%) & 2/10 (20.0\%) & 0/0 & 900\\
\midrule
Episode-micro total & 31/96 (32.3\%) & 25/86 (29.1\%) & 4/14 & --\\
Task-macro average & 32.0\% & 28.0\% & -- & --\\
\bottomrule
\end{tabular*}
\end{table*}

\noindent\textbf{Trained-policy references.}
ACT, DP, DP3, RDT, and $\pi_0$ are published results from RoboTwin~\citep{robotwin2}, Appendix K, Table~10, averaged over the same ten task names. The source trains on 50 clean demonstrations per task and evaluates 100 rollouts. Here $\pi_0$ is distinct from the $\pi_{0.5}$ used in our other experiments.

\begin{table*}[!t]
\centering\small
\setlength{\tabcolsep}{5pt}
\caption[Published RoboTwin baseline accuracies]{\textbf{Published RoboTwin task accuracies (\%).} Each cell is Easy/Hard. Averages are recomputed over the ten displayed tasks from RoboTwin~\citep{robotwin2}, Appendix K, Table~10.}
\label{tab:robotwin_baselines}
\begin{tabular*}{\linewidth}{@{\extracolsep{\fill}}lrrrrr@{}}
\toprule
Task & ACT & DP & DP3 & RDT & $\pi_0$\\
\midrule
Grab Roller & 94/25 & \textbf{98}/0 & \textbf{98}/2 & 74/43 & 96/\textbf{80}\\
Handover Mic & 85/0 & 53/0 & \textbf{100}/3 & 90/\textbf{31} & 98/13\\
Lift Pot & 88/0 & 39/0 & \textbf{97}/0 & 72/9 & 84/\textbf{36}\\
Move Can Pot & 22/4 & 39/0 & \textbf{70}/6 & 25/12 & 58/\textbf{21}\\
Place Dual Shoes & 9/0 & 8/0 & 13/0 & 4/\textbf{4} & \textbf{15}/0\\
Place Phone Stand & 2/0 & 13/0 & \textbf{44}/2 & 15/6 & 35/\textbf{7}\\
Put Bottles Dustbin & 27/1 & 22/0 & \textbf{60}/\textbf{21} & 21/4 & 54/13\\
Put Object Cabinet & 15/0 & 42/0 & \textbf{72}/1 & 33/\textbf{18} & 68/\textbf{18}\\
Stack Blocks Two & 25/0 & 7/0 & 24/0 & 21/\textbf{2} & \textbf{42}/1\\
Stack Bowls Two & 82/0 & 61/0 & 83/6 & 76/30 & \textbf{91}/\textbf{41}\\
\midrule
Task-macro average & 44.9/3.0 & 38.2/0.0 & \textbf{66.1}/4.1 & 43.1/15.9 & 64.1/\textbf{23.0}\\
\bottomrule
\end{tabular*}
\end{table*}

\noindent\textbf{Interpretation.}
K1 loses 4.0 percentage points from Easy to Hard, compared with 27.2 for RDT and 41.1 for $\pi_0$ on this task subset. Re-measuring geometry helps the agent preserve performance under changes in clutter, appearance, and table layout. Strong roller and block-stacking results coexist with failures on shoes and bottle disposal, locating the remaining bottleneck in long-horizon bimanual coordination.

\subsection{RoboCasa: mobile-base and torso control}
\label{app:robocasa}
The same Gemini agent controls a PandaOmron arm, gripper, planar base, base rotation, and torso height on 24 kitchen tasks. Ten initial states per task yield 240 episodes, evaluated with native completion feedback. K1 succeeds in 53 episodes (22.1\%), spanning all six task categories (Table~\ref{tab:robocasa_detail}). Counter-to-sink placement reaches 80.0\%, drawer closing 60.0\%, and cabinet-to-counter placement and faucet-spout adjustment each reach 50.0\%.

Results depend strongly on the direction of an interaction. Cabinet-to-counter placement succeeds in five trials, while counter-to-cabinet placement remains unsolved. Turning off the microwave succeeds in five trials, while activation remains unsolved. These contrasts expose difficulty with constrained destinations and appliance contact even when the interface supports the required motions. Coffee preparation is the weakest category: mug serving succeeds twice, while mug setup and button pressing remain unsolved. Table~\ref{tab:robocasa_tasks} gives all 24 task results.

\begin{table*}[!t]
\centering\small
\setlength{\tabcolsep}{5pt}
\renewcommand{\arraystretch}{1.12}
\caption[RoboCasa mobile manipulation]{\textbf{Mobile manipulation results on RoboCasa.} K1 uses Gemini 3.7 Flash with arm, gripper, mobile-base, and torso control, without target-environment fine-tuning. Each task is evaluated on 10 initial states.}
\label{tab:robocasa_detail}
\begin{tabularx}{\linewidth}{@{}Yrrr@{}}
\toprule
Task category & Tasks & Successful episodes & Success rate\\
\midrule
Pick-and-place & 8 & 18/80 & 22.5\%\\
Doors and drawers & 6 & 14/60 & 23.3\%\\
Stove control & 2 & 4/20 & 20.0\%\\
Faucet control & 3 & 10/30 & 33.3\%\\
Microwave operation & 2 & 5/20 & 25.0\%\\
Coffee preparation & 3 & 2/30 & 6.7\%\\
\midrule
\rowcolor{kpale}\textbf{Overall} & \textbf{24} & \textbf{53/240} & \textbf{22.1\%}\\
\bottomrule
\end{tabularx}
\end{table*}
\begin{table*}[!t]
\centering\small
\setlength{\tabcolsep}{5pt}
\renewcommand{\arraystretch}{1.12}
\caption[RoboCasa per-task results]{\textbf{Per-task RoboCasa results.} K1 with Gemini 3.7 Flash is evaluated on 10 initial states per task \mbox{without target-environment fine-tuning}.}
\label{tab:robocasa_tasks}
\begin{tabularx}{\linewidth}{@{}lYrr@{}}
\toprule
Category & Task & Successful episodes & Success rate\\
\midrule
Pick-and-place & Counter $\rightarrow$ cabinet & 0/10 & 0\%\\
 & Cabinet $\rightarrow$ counter & 5/10 & 50\%\\
 & Counter $\rightarrow$ sink & 8/10 & 80\%\\
 & Sink $\rightarrow$ counter & 3/10 & 30\%\\
 & Counter $\rightarrow$ microwave & 0/10 & 0\%\\
 & Microwave $\rightarrow$ counter & 0/10 & 0\%\\
 & Counter $\rightarrow$ stove & 1/10 & 10\%\\
 & Stove $\rightarrow$ counter & 1/10 & 10\%\\
\midrule
Doors and drawers & Open single door & 1/10 & 10\%\\
 & Close single door & 3/10 & 30\%\\
 & Open double door & 0/10 & 0\%\\
 & Close double door & 0/10 & 0\%\\
 & Open drawer & 4/10 & 40\%\\
 & Close drawer & 6/10 & 60\%\\
\midrule
Stove control & Turn on stove & 4/10 & 40\%\\
 & Turn off stove & 0/10 & 0\%\\
\midrule
Faucet control & Turn on sink faucet & 1/10 & 10\%\\
 & Turn off sink faucet & 4/10 & 40\%\\
 & Turn sink spout & 5/10 & 50\%\\
\midrule
Microwave operation & Turn on microwave & 0/10 & 0\%\\
 & Turn off microwave & 5/10 & 50\%\\
\midrule
Coffee preparation & Set up mug & 0/10 & 0\%\\
 & Serve mug & 2/10 & 20\%\\
 & Press coffee-machine button & 0/10 & 0\%\\
\midrule
\rowcolor{kpale}\textbf{Overall} & \textbf{24 tasks} & \textbf{53/240} & \textbf{22.1\%}\\
\bottomrule
\end{tabularx}
\end{table*}

\subsection{Real-robot protocol and execution frames}
\label{app:real_robot}
We evaluate a physical Piper arm with a parallel-jaw gripper and two external RGB-D cameras. Both methods use Gemini 3.7 Flash through the same Piper control and inverse-kinematics adapters. The Clean baseline follows uni-robot-agent~\citep{uniagent}, receives RGB observations and measured arm pose, and exposes four tools: \tool{capture\_scene}, \tool{move}, \tool{hold}, and \tool{finish}. K1 adds perception tools and memory.\looseness=-1

Three concluded trials per method and task give nine scored trials for each method. Outcomes are assessed through operator confirmation and image review. For a single object, approach, grasp-and-lift, transport, and stable placement each contribute 0.5 progress points, for a maximum of 2. Can placement is judged by containment. Main-paper Table~\ref*{M-tab:real_robot} reports the success counts and progress scores.\looseness=-1

\begin{figure*}[!t]
\centering
\includegraphics[width=\textwidth]{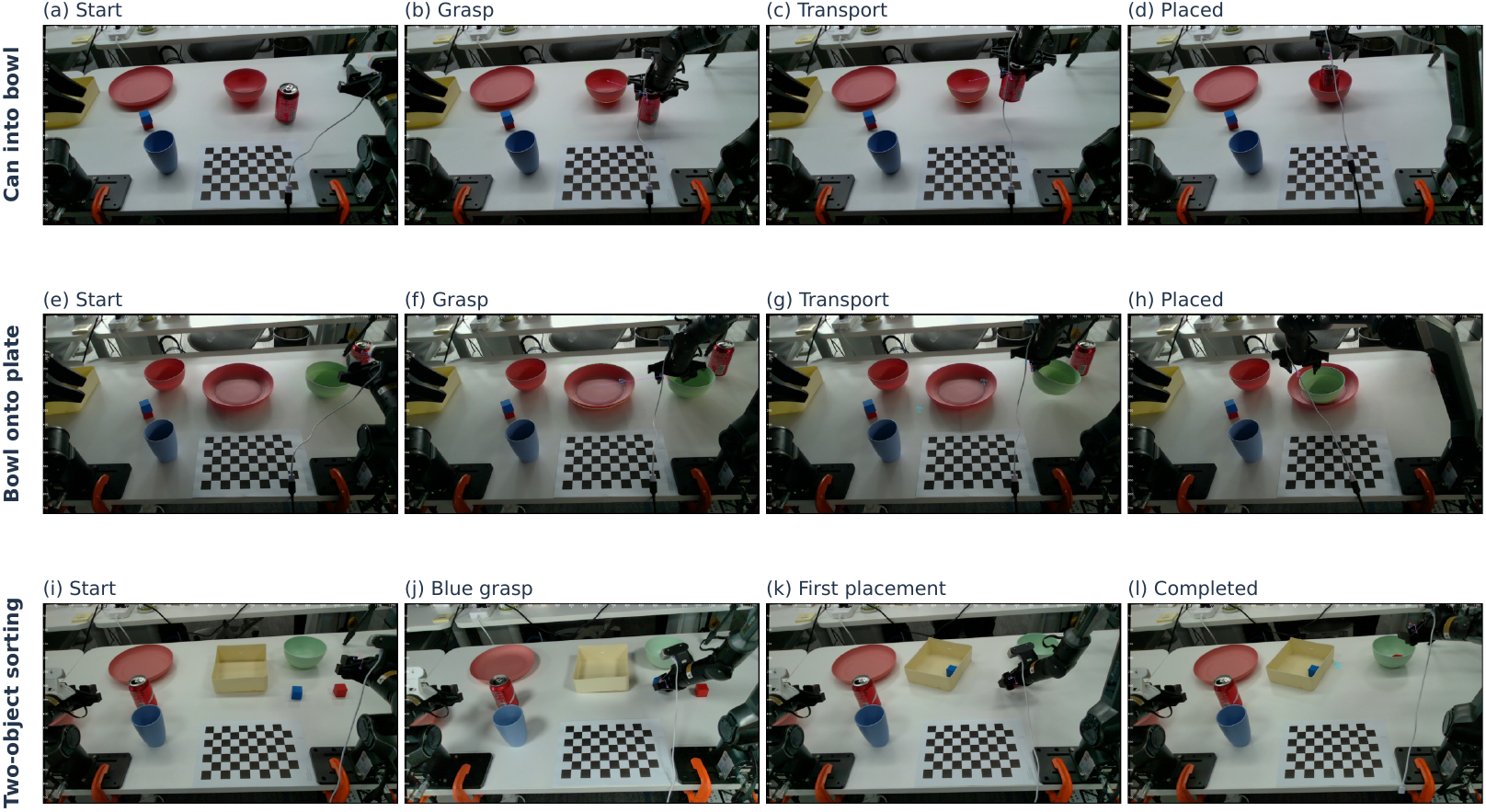}
\caption[Physical-robot video frames]{\textbf{Physical-robot execution across all three tasks.} Rows show successful K1 trials for can placement, bowl placement, and two-object sorting. The head-camera frames follow the sequence from the initial scene through grasping and transport to the completed placement.}
\label{fig:real_robot_all_tasks}
\end{figure*}

Both methods use a 50-call limit for can and bowl placement and a 100-call limit for sorting. Figure~\ref{fig:real_robot_all_tasks} shows head-camera frames from successful K1 trials on all three tasks.

\subsection{Mixed-environment post-training}
\label{app:mixed_posttraining}
We continue agentic SFT of the LIBERO-trained Qwen3.5-9B agent from its epoch-3 checkpoint on 316 successful trajectories: 107 from LIBERO-PRO, 142 from RoboSuite, 32 from RoboCasa, and 35 from RoboTwin. The same tool-call objective updates the language adapter for three additional epochs. Table~\ref{tab:mixed_posttraining_detail} reports the paired before/after evaluation used in the main paper.

Continuation improves LIBERO-PRO new-state accuracy from 42.9\% to 52.4\%, held-out-task accuracy from 11.1\% to 27.8\%, and RoboSuite accuracy from 28.3\% to 50.0\%. RoboCasa remains low and RoboTwin remains near 2.5\%, showing that the expanded data strengthens established single-arm skills before it unlocks mobile or bimanual control.
\begin{table*}[!t]
\centering\small
\setlength{\tabcolsep}{5pt}
\renewcommand{\arraystretch}{1.12}
\caption[Mixed-environment post-training]{\textbf{Performance before and after mixed-environment post-training.} LIBERO-PRO measures source-domain generalization, and the other environments use their held-out evaluation sets. Each environment retains the same evaluation interface and completion protocol before and after training. Counts denote successful/evaluated episodes, and gains are percentage points.}
\label{tab:mixed_posttraining_detail}
\begin{tabularx}{\linewidth}{@{}Yrrr@{}}
\toprule
Evaluation & Before & After & Gain\\
\midrule
\multicolumn{4}{@{}l}{\emph{Source-domain generalization}}\\
LIBERO-PRO B: new initial states & 18/42 (42.9\%) & \textbf{22/42 (52.4\%)} & \textbf{+9.5}\\
LIBERO-PRO C: held-out task conditions & 4/36 (11.1\%) & \textbf{10/36 (27.8\%)} & \textbf{+16.7}\\
\midrule
\multicolumn{4}{@{}l}{\emph{Additional environments}}\\
\rowcolor{kpale}RoboSuite & 17/60 (28.3\%) & \textbf{30/60 (50.0\%)} & \textbf{+21.7}\\
RoboCasa & 4/48 (8.3\%) & 2/48 (4.2\%) & $-4.2$\\
RoboTwin & 1/39 (2.6\%) & 1/40 (2.5\%) & ---\\
\bottomrule
\end{tabularx}
\end{table*}

\section{Additional Analysis}

\subsection{Perception and completion-feedback ablations}
\label{app:po_ablation}
Each ablation uses 30 matched episodes per environment. LIBERO-PRO covers Spatial, Object, and Goal tasks with swap and task variants. RoboTwin covers ten tasks across clean and randomized conditions.

Disabling perception retains RGB observations, robot state, motion feedback, and memory. Harness completion feedback permits continued execution after an unsuccessful completion request and stops on native success. RoboTwin retains native environment termination in all conditions.

\begin{table}[!t]
\centering\small
\caption{\textbf{Perception and completion feedback (\%).} Thirty matched episodes per environment. RoboTwin always retains \mbox{native environment termination}.}
\label{tab:po_ablation}
\begin{tabular*}{\linewidth}{@{\extracolsep{\fill}}ccrr@{}}
\toprule
Perception & Feedback & LIBERO-PRO & RoboTwin\\
\midrule
Off & Off & 10.0 & 0.0\\
On & Off & 30.0 & 6.7\\
Off & On & 13.3 & 6.7\\
On & On & \textbf{70.0} & \textbf{30.0}\\
\bottomrule
\end{tabular*}
\end{table}

The combined configuration performs best in both environments (Table~\ref{tab:po_ablation}). Perception raises success from 10.0\% to 30.0\% on LIBERO-PRO and from 0.0\% to 6.7\% on RoboTwin without completion feedback. With feedback enabled, the full system reaches 70.0\% and 30.0\%, showing that measured evidence and iterative correction reinforce each other.

\subsection{Perception-tool configuration details}
\label{app:tool_ablation}
The main-paper tool-family comparison evaluates the full perception stack and configurations with one family removed. Each configuration uses the same 30 episodes per environment. Removing a family also updates the available-tool instructions. Table~\ref{tab:tool_ablation} reports exact counts for the configurations summarized in the main paper.

\begin{table*}[!t]
\centering\small
\caption{\textbf{Perception-tool family ablations.} Counts give successful/evaluated episodes and percentages.}
\label{tab:tool_ablation}
\begin{tabularx}{\linewidth}{@{}Yrr@{}}
\toprule
Configuration & LIBERO-PRO & RoboTwin\\
\midrule
Full perception tools & \textbf{21/30 (70.0\%)} & \textbf{9/30 (30.0\%)}\\
Without point-depth/local geometry tools & 17/30 (56.7\%) & 5/30 (16.7\%)\\
Without region tools and their dependents & 15/30 (50.0\%) & 4/30 (13.3\%)\\
Without cross-frame tracking & 12/30 (40.0\%) & 6/30 (20.0\%)\\
Without grasp candidates & 18/30 (60.0\%) & 4/30 (13.3\%)\\
\bottomrule
\end{tabularx}
\end{table*}

The complete perception stack performs best in both environments. Tracking removal has the largest effect on LIBERO-PRO, while region or grasp removal has the largest effect on RoboTwin. The tracking condition removes both explicit tracking calls and the automatic anchor refresh that keeps measured points current after motion. Since the model never requests the explicit tracking tools in these trajectories, the LIBERO-PRO drop identifies persistent anchor refresh as the useful component. Bimanual RoboTwin tasks depend more strongly on grounded regions and candidate hand poses.

\subsection{Tool-use distribution}
\label{app:toolusage}
The tool-use analysis covers the complete 180-episode Gemini evaluation on LIBERO-PRO: Spatial, Object, and Goal, each with ten task indices, two variants, and three initial states. This gives 30 base task indices and 60 task-variant conditions, with 139 successful and 41 unsuccessful trajectories. RoboSuite, RoboTwin, Astra, and student rollouts are excluded from this count.

All 6,282 model responses are included, averaging 34.9 per episode. Main-paper Figure~\ref*{M-fig:toolusage} partitions them by function. Motion accounts for 4,238 decisions (67.5\%), perception for 1,604 (25.5\%), memory/progress for 245 (3.9\%), completion for 142 (2.3\%), and output-limit responses for the remaining 0.8\%.

Perception and memory/progress contribute 1,849 non-motion decisions (29.4\%). Of the perception requests, 1,036 (64.6\%) occur after the initial frame across 159 trajectories. The agent therefore gathers evidence throughout execution rather than only during scene initialization.

\begin{table}[!t]
\centering\small
\setlength{\tabcolsep}{3pt}
\caption[Individual tool-request counts]{Tool-request counts for Figure~\ref*{M-fig:toolusage}, including \mbox{requests that return errors}.}
\label{tab:toolusage}
\begin{tabularx}{\linewidth}{@{}lYr@{}}

\toprule
Category & Tool & Requests\\
\midrule
Motion & \tool{move\_toward} & 1,835\\
 & \tool{move\_relative} & 654\\
 & \tool{move\_to\_pose} & 547\\
 & \tool{rotate\_toward} & 271\\
 & \tool{align\_axis} & 48\\
 & \tool{set\_gripper} & 883\\
\midrule
Perception & \tool{measure\_depth} & 733\\
 & \tool{find\_regions} & 384\\
 & \tool{inspect\_region} & 187\\
 & \tool{inspect\_crop} & 13\\
 & \tool{grasp\_candidates} & 274\\
 & \tool{align\_region\_references} & 13\\
\midrule
Memory & \tool{update\_task\_progress} & 121\\
 & \tool{search\_history} & 66\\
 & \tool{read\_history} & 58\\
Completion & \tool{done} & 142\\
\bottomrule
\end{tabularx}
\end{table}

\tool{align\_region\_references} and grasp proposals are counted as perception because they return geometric evidence without moving the robot. Persistent visual anchors refresh between calls, which explains why explicit tracking requests do not appear in Table~\ref{tab:toolusage}.

\subsection{Memory-context ablation}
\label{app:memory_ablation}

$K$ controls how many previous image rounds accompany the current observation, and $N$ controls the recent text window. We vary $K\in\{0,1,2,3,4\}$ at $N=8$ and $N\in\{1,2,4,8,16\}$ at $K=1$. Retrieval, progress memory, perception tools, motion tools, and completion feedback remain enabled. Each setting uses the same 18 cases spanning Spatial, Object, and Goal with swap and task variants.

\begin{table}[!t]
\centering\small
\caption[Memory-context ablation results]{\textbf{Memory-context ablation outcomes.} Each row uses the same 18 cases. \shortstack{Mean\\calls} include successes and failures. Both evaluations share $K=1,N=8$, counted once.}
\label{tab:memory_ablation}
\begin{tabular*}{\linewidth}{@{\extracolsep{\fill}}rrrrr@{}}
\toprule
$K$ & $N$ & Successes & Accuracy (\%) & \shortstack{Mean\\calls}\\
\midrule
0 & 8 & 15/18 & 83.3 & 28.3\\
1 & 8 & 14/18 & 77.8 & 33.0\\
2 & 8 & 14/18 & 77.8 & 41.2\\
3 & 8 & 11/18 & 61.1 & 47.4\\
4 & 8 & 12/18 & 66.7 & 45.9\\
\midrule
1 & 1 & 13/18 & 72.2 & 70.0\\
1 & 2 & 13/18 & 72.2 & 72.6\\
1 & 4 & 10/18 & 55.6 & 50.7\\
1 & 16 & 14/18 & 77.8 & 36.0\\
\bottomrule
\end{tabular*}
\end{table}

\noindent\textbf{Observed effect of image history.}
At $N=8$, $K=0$ reaches 83.3\%, $K=1$ and $K=2$ reach 77.8\%, and longer histories reach 61.1--66.7\%. Mean calls also increase as more images are included. Current observations plus retrieval are more effective than automatically carrying several stale views.

\noindent\textbf{Observed effect of text history.}
At $K=1$, $N=8$ and $N=16$ reach 77.8\%, while $N=1$ and $N=2$ reach 72.2\% and require more than twice as many calls. A moderate text window preserves recent measurements and action outcomes, reducing repeated queries.

\noindent\textbf{Practical implication.}
Text history carries decisions and milestones, while images provide time-sensitive evidence. The default $K=1,N=8$ retains one prior visual comparison and enough recent text to maintain continuity.

\subsection{Runtime efficiency}
\label{app:runtime_efficiency}

We measure runtime from the recorded LIBERO-PRO evaluations (Table~\ref{tab:runtime_efficiency}). Gemini with K1 takes an average of 5.16 minutes per episode across 180 episodes, and 3.42 minutes among successful episodes. On the 60 matched configurations, Astra with K1 achieves 93.3\% success with an average runtime of 7.31 minutes, compared with 63.3\% success and 4.47 minutes for RGB-only control. K1 supports more rounds of observation and correction, trading additional interaction time for higher task completion.

\begin{table*}[!t]
\centering\small
\setlength{\tabcolsep}{4pt}
\caption[Runtime efficiency on LIBERO-PRO]{\textbf{Runtime statistics on LIBERO-PRO.} Episode-level statistics include successful and unsuccessful outcomes unless indicated. The two Astra configurations use the same 60 cases, and Gemini statistics cover the full 180-case evaluation. Mean request time is computed over recorded model calls. Episode times are in minutes and request times in seconds.}
\label{tab:runtime_efficiency}
\begin{tabular*}{\linewidth}{@{\extracolsep{\fill}}lrrrrrr@{}}
\toprule
Agent & Episodes & Success & \shortstack{Episode time\\mean / median} & \shortstack{Success-only\\mean time} & \shortstack{Mean\\calls} & \shortstack{Mean\\request time}\\
\midrule
Gemini + K1 & 180 & 77.2\% & 5.16 / 3.59 & 3.42 & 34.9 & 7.07\\
Astra RGB-only & 60 & 63.3\% & 4.47 / 3.70 & 3.53 & 24.8 & 10.33\\
Astra + K1 & 60 & 93.3\% & 7.31 / 5.54 & 6.61 & 45.9 & 8.11\\
\bottomrule
\end{tabular*}
\end{table*}

Model requests account for 77\% of episode time for Gemini with K1 and 83\% for Astra with K1. Serving latency and the number of decision rounds therefore dominate the runtime cost.

\subsection{Gemini RGB-only failure analysis}
\label{app:gemini_rgb}
Gemini with the RGB-only interface succeeds on 10 of the 60 matched configurations: 1/20 Spatial, 8/20 Object, and 1/20 Goal (Table~\ref{tab:paired}). Episodes average 44.8 model calls and 4.92 minutes. Of the 50 failures, 25 exhaust the 1,200-step budget, 17 end when the model submits \tool{done} on an unsolved task, and eight end with \tool{give\_up}. The RGB-only interface provides no completion feedback, so an incorrect \tool{done} ends the episode.

One episode illustrates the failure. In Spatial swap task 0 at state 0, the decision notes name the instructed bowl between the plate and the ramekin and the plate as its destination. After releasing the bowl, the agent keeps the gripper at one height and issues six consecutive \tool{move\_to} calls (calls 21 to 26) whose lateral targets range from $y=-0.10$\,m to $y=0.04$\,m. Each note describes the motion as centering over the bowl. The agent closes the gripper repeatedly and reaches the step budget after 64 calls without a valid placement. The semantic targets are correct, while the 3D coordinates inferred from the two RGB views are not, which motivates returning measured spatial relations through tools.

\subsection{Qualitative interaction traces}
\label{app:trace}

\noindent\textbf{Where K1 helps: spatially specified manipulation.}
The matched 60-episode Astra evaluation shows the largest absolute gain on Spatial tasks: success rises from 45\% with RGB-only control to 90\% with K1, a 45-point increase. The corresponding gains are 20 points on Object and 25 points on Goal. This ordering holds for both perturbation variants (Table~\ref{tab:spatial_gain_breakdown}). Gemini shows the same ordering, with gains of 70, 55, and 40 points on Spatial, Goal, and Object (Table~\ref{tab:paired}). Spatial tasks require selecting an object through its relation to nearby objects and aligning it with a destination. K1 supports these decisions with grounded regions and measured geometric relations. The paired trajectory below illustrates how the agent uses this evidence during execution. These category-level gains describe the full K1 system, whereas the perception ablations in Section~\ref{app:tool_ablation} isolate the contribution of its perception tools.

\begin{table*}[!t]
\centering\small
\setlength{\tabcolsep}{5pt}
\caption[Category-level gains in the matched Astra evaluation]{\textbf{Spatial tasks show the largest absolute gain in both variants.} Each variant contains ten paired episodes per category. Entries give RGB-only and K1 successes, and gains are percentage points.}
\label{tab:spatial_gain_breakdown}
\begin{tabularx}{\linewidth}{@{}lYYr@{}}
\toprule
Category & Swap: RGB $\rightarrow$ K1 & Task: RGB $\rightarrow$ K1 & Overall gain\\
\midrule
Spatial & $5/10 \rightarrow 9/10$ (+40) & $4/10 \rightarrow 9/10$ (+50) & +45\\
Object & $8/10 \rightarrow 10/10$ (+20) & $8/10 \rightarrow 10/10$ (+20) & +20\\
Goal & $7/10 \rightarrow 9/10$ (+20) & $6/10 \rightarrow 9/10$ (+30) & +25\\
\bottomrule
\end{tabularx}
\end{table*}

\noindent\textbf{Paired gain case: grounded lift and placement.}
\begin{figure*}[!t]
\centering
\includegraphics[width=.90\linewidth]{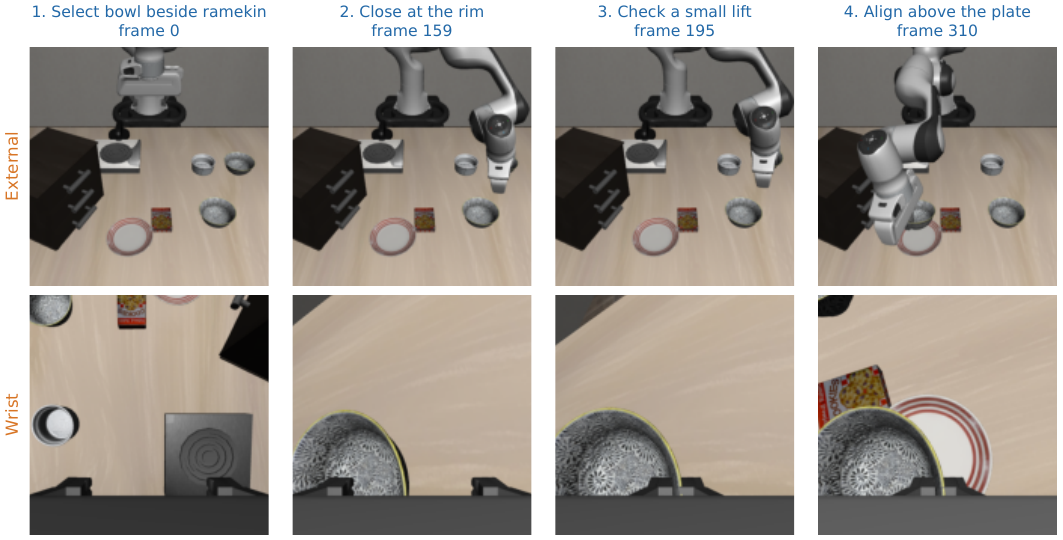}
\caption[Recorded paired-case trajectory]{\textbf{Recorded LIBERO-PRO trajectory.} GPT-6 Astra with K1 on Spatial swap task 1/state 0. Panels show original external/wrist RGB at four decision frames, without candidate or region overlays. The final panel precedes native termination. Table~\ref{tab:trace} summarizes \mbox{the tool evidence}.}
\label{fig:trace}
\end{figure*}

The instruction is to pick the black bowl next to the ramekin and place it on the plate. Both Astra episodes start from the same state. The RGB-only agent attempts the transfer, opens its gripper, and submits \tool{done} after 13 calls and 235 steps, and the episode is unsuccessful. Its own final record identifies a placement problem: the controller stayed near a height of 1.06\,m despite lower commanded targets. Thus, the agent noticed execution shortfall but ended the attempt before resolving it.

With K1, the agent selects bowl region S2 using the ramekin relation, refreshes it from the wrist view, and checks carrying through a small lift. It then selects plate region S3, queries source-to-destination alignment, and makes local corrections (Table~\ref{tab:trace}). The episode succeeds after 33 calls and 322 steps and passes the evaluator's post-release check. The useful pattern is an explicit sequence of grounded target selection, physical verification, and measured correction: each stage supplies evidence for the next action.

\begin{table*}[!t]
\centering\small
\caption[Selected tool calls from the K1 trajectory]{Selected calls from the LIBERO-PRO trace. Intermediate calls adjust poses or retrieve history within the same episode.}
\label{tab:trace}
\begin{tabularx}{\linewidth}{@{}r>{\raggedright\arraybackslash}p{.24\linewidth}Y@{}}
\toprule
Call & Operation & Observable role\\
\midrule
0 & \tool{find\_regions} & Search for black bowls. S2 is selected from the existing external view based on the ramekin relation.\\
2 & \tool{grasp\_candidates} & Request nominal grasp hypotheses for S2.\\
3--9 & Transit and rotation & Approach with open jaws, rotate in bounded steps, and begin the persistent pose approach.\\
10, 17 & \tool{inspect\_region} & Refresh S2 using a wrist-image bounding box while approaching the rim.\\
20 & \tool{set\_gripper} & Close at frame 159 and allow native settling.\\
21--23 & Inspect, lift, update & Reinspect S2, request a 25\,mm lift, and update the carry milestone from the resulting observation.\\
25--28 & Find plate and align & Select destination S3 and query a horizontal source--destination reference alignment before transit.\\
29--32 & Local corrections & Reinspect the measured relation and refine the position in bounded steps. Native success ends the rollout at frame 322.\\
\bottomrule
\end{tabularx}
\end{table*}

\clearpage
\noindent\textbf{Remaining bottleneck: turning measured error into effective recovery.}
In a second episode of Astra with K1, the task is to place a plate on top of the drawer (Goal task 4, initial state 0). At step 1,120, the agent records that it is still adjusting the plate orientation. The next corrections produce little motion, and K1 reports the execution shortfall explicitly. A subsequent alignment query also returns a sizable remaining horizontal error (Table~\ref{tab:failure_cases}). The agent continues adjusting until the 1,200-step budget is exhausted.

\begin{table}[H]
\centering\small
\setlength{\tabcolsep}{5pt}
\caption[Measured error during an unsuccessful placement]{\textbf{A feedback-to-recovery gap.} Selected records from the unsuccessful plate-placement episode. Call indices are zero-based.}
\label{tab:failure_cases}
\begin{tabularx}{\linewidth}{@{}lY@{}}
\toprule
Call & Evidence available to the agent\\
\midrule
87 & The agent's progress record identifies orientation adjustment as the current stage.\\
92--93 & Motion feedback reports undertracking: achieved axial progress is approximately 26\% and 10\% of the requested motion.\\
95 & Region alignment reports a remaining horizontal reference error of 6.7\,cm at step 1,192.\\
96 & Another pose correction consumes the remaining execution budget. The episode ends unsuccessfully after 97 calls.\\
\bottomrule
\end{tabularx}
\end{table}

K1 exposes the execution shortfall and residual placement error, but the agent repeats local corrections instead of changing its approach or regrasping. This case identifies recovery planning as the next requirement beyond spatial measurement.

\begin{figure}[H]
\centering
\includegraphics[width=.80\linewidth]{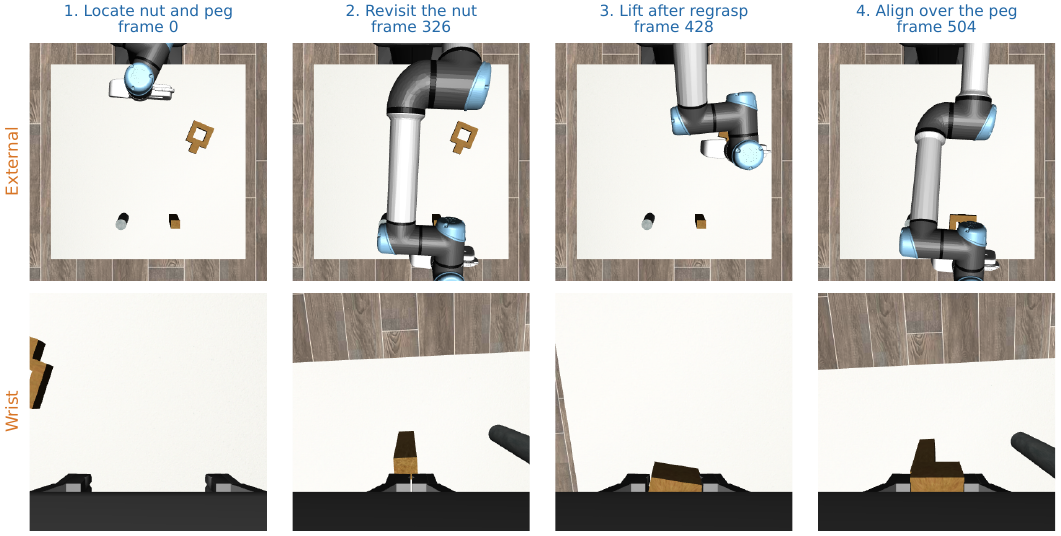}
\caption[RoboSuite UR5e nut-assembly trajectory]{\textbf{RoboSuite nut assembly with regrasping.} Recorded external and wrist RGB at four decision frames. The nut remains on the table at frame 326. After the third closure, it is visible near the gripper at frame 428 and above the peg at frame 504. The last panel precedes the final descent and native success at frame 534.}
\label{fig:robosuite_trace}
\end{figure}

\clearpage
\noindent\textbf{RoboSuite: cross-embodiment nut assembly with regrasping.}
\label{app:robosuite_trace}
Gemini with K1 controls a UR5e arm with the Panda gripper to place a square nut onto its peg. The episode reaches success after 35 model calls and 534 control steps. Figure~\ref{fig:robosuite_trace} and Table~\ref{tab:robosuite_trace} show the intermediate regrasp attempts and visual checks of nut carrying.\looseness=-1

\begin{table}[H]
\centering\small
\caption[Selected calls from the RoboSuite trajectory]{Selected calls from the 35-call RoboSuite episode. Omitted calls perform measurements or local motion within the same episode.}
\label{tab:robosuite_trace}
\begin{tabularx}{\linewidth}{@{}r>{\raggedright\arraybackslash}p{.24\linewidth}Y@{}}
\toprule
Call & Operation & Observable role\\
\midrule
0--3 & Ground and inspect & Find the nut, refine its handle as S2, measure depth, and request grasp hypotheses.\\
4--9 & Approach and close & Record progress, approach using wrist depth, close at frame 79, and request a 3\,cm lift.\\
10--14 & Measure and revisit & Measure the scene and move toward the peg. A new search at frame 171 finds the nut still near its original position.\\
15--23 & Second attempt & Return with open jaws, remeasure, close at frame 235, and repeat lift and transport.\\
24--29 & Reground and regrasp & Find the nut again at frame 326, use wrist depth to refine approach, and close at frame 384.\\
30--31 & Lift and inspect & Lift with closed jaws. The frame-428 wrist image shows the nut near the gripper; a depth query follows.\\
32--34 & Align and descend & Move above the peg, query wrist depth, then descend. Execution stops on native success at frame 534.\\
\bottomrule
\end{tabularx}
\end{table}

The episode illustrates recovery through renewed visual grounding: after unsuccessful transport attempts, the agent locates the nut on the table and updates its grasp approach. The same perception-action loop operates through the UR5e adapter without target-task fine-tuning.

\begin{figure}[H]
\centering
\includegraphics[width=.80\linewidth]{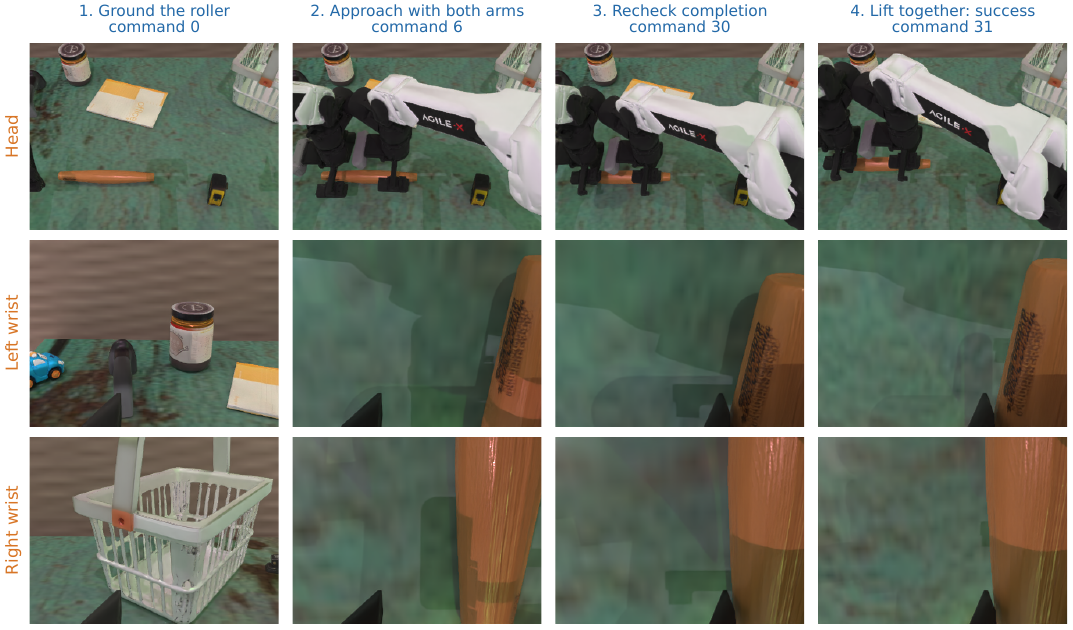}
\caption[RoboTwin Hard dual-arm roller trajectory]{\textbf{Dual-arm grasping in a randomized RoboTwin scene.} Recorded head, left-wrist, and right-wrist RGB. Command indices count native action requests. The final panel shows the terminal observation after both arms lift.}
\label{fig:robotwin_trace}
\end{figure}

\clearpage
\noindent\textbf{RoboTwin: dual-arm grasping under visual perturbation.}
\label{app:robotwin_trace}
In the Hard condition of RoboTwin \texttt{grab\_roller} (seed 10002), Gemini with K1 follows the instruction ``Hold the smooth wooden roller with both arms.'' The Aloha AgileX episode succeeds after 12 model calls and 31 native action requests. The sequence includes a rejected completion claim followed by a successful joint lift (Figure~\ref{fig:robotwin_trace} and Table~\ref{tab:robotwin_trace}).

\begin{table}[H]
\centering\small
\caption[Selected calls from the RoboTwin trajectory]{Calls from the 12-call RoboTwin episode, including the failed target and rejected completion check.}
\label{tab:robotwin_trace}
\begin{tabularx}{\linewidth}{@{}r>{\raggedright\arraybackslash}p{.24\linewidth}Y@{}}
\toprule
Call & Operation & Observable role\\
\midrule
0--2 & Ground and propose & Find roller region S1 in the head view, then request grasp hypotheses separately for the left and right arms.\\
3--5 & \tool{move\_effectors} & Approach, lower, and close both grippers using paired arm targets.\\
6--7 & Adjust right grasp & Reopen and reposition the right gripper, then close it while retaining the left grasp command.\\
8--9 & Correct a failed target & A negative left-arm height produces planner failures. The next call restores a positive height and refines the right grasp.\\
10 & \tool{done} & Completion check fails; execution continues.\\
11 & \tool{move\_effectors} & Lift both closed grippers by approximately 6\,cm in achieved motion. Native success follows at command 31.\\
\bottomrule
\end{tabularx}
\end{table}

This case shows how shared visual grounding supports arm-specific targets and coordinated motion. Sparse completion feedback also separates an agent's verbal claim from physical success, allowing the remaining action to finish the task.\looseness=-1